\documentclass[lettersize,journal]{IEEEtran}

\usepackage{graphics} 
\usepackage{epsfig} 
\usepackage{times}
\usepackage{graphicx}
\usepackage{float}
\usepackage{times} 
\usepackage{amsmath} 
\usepackage{amssymb}  
\usepackage{booktabs}
\usepackage{epstopdf}
\usepackage{color}
\usepackage{url}
\usepackage{cite}
\usepackage{balance}
\usepackage{textcomp}
\usepackage{makecell}
\usepackage{threeparttable}

\usepackage{color}
\usepackage{xcolor}
\usepackage{multirow}
\usepackage{colortbl}
\usepackage{picinpar}
\usepackage{stfloats}
\usepackage{url}

\usepackage{todonotes}

\usepackage{colortbl}
\usepackage{float} 

\usepackage{mathtools}
\usepackage{listings}                 
\usepackage{balance}
\usepackage{ragged2e} 
\usepackage[switch]{lineno}
\modulolinenumbers[5]

\usepackage[pagebackref=true,breaklinks=true,colorlinks,bookmarks=false]{hyperref}
\usepackage{subcaption}

\usepackage[noabbrev,nameinlink,capitalise]{cleveref} 
\Crefformat{figure}{#2Fig.~#1#3}
\Crefmultiformat{figure}{Figs.~#2#1#3}{ and~#2#1#3}{, #2#1#3}{ and~#2#1#3}

\newcommand{\mb}[1]{\mathbf{#1}}

\hypersetup{
    colorlinks=true, 
    linkcolor=red, 
    citecolor=green, 
    urlcolor=red, 
    linkbordercolor=red, 
    citebordercolor=green, 
    urlbordercolor=red 
}

\definecolor{lightgreen}{RGB}{220, 245, 220}
\newcommand{\rev}[1]{{\color{black}#1}}
\newcommand{\revision}[1]{\textcolor{black}{#1}}

\begin{document}

\title{Accurate Measurement of 3D and 2D Circular Centers With Application to LiDAR-Camera Extrinsic Calibration}

\author{Jiajun~Jiang, Xiao Hu$^{*}$, Wancheng~Liu, and Wei Jiang
\thanks{This work was supported in part by the National Natural Science Foundation of China under Grant T2422002. (Corresponding author: Xiao Hu.)}
\thanks{Jiajun~Jiang is with the Thrust of Robotics and Autonomous
Systems, The Hong Kong University of Science and Technology (Guangzhou), 511453, China (email: jjiang127@connect.hkust-gz.edu.cn).}
\thanks{Xiao Hu is with Astrall dynamics Technology, ShenZhen, 510085, China(e-mail: hux062303@gmail.com).}
\thanks{Wancheng Liu is with the Horizon-Continental Technology Corporation, Shanghai, 201800, China (email: wancheng.liu@horizon-continental.com). }
\thanks{Wei Jiang is with the State Key Laboratory of Rail Traffic Control and Safety and the Beijing Engineering Research Center of EMC and GNSS Technology for Rail Transportation, School of Electronic and Information Engineering, Beijing Jiaotong University, Beijing 100044, China (e-mail: weijiang@bjtu.edu.cn).}
}

\markboth{IEEE Transactions on Instrumentation and Measurement}%
{Jiang \MakeLowercase{\textit{et al.}}: Accurate Measurement of 3D and 2D Circular Centers}


\maketitle

\begin{abstract}
Accurate measurement of circular centers is a fundamental geometric sensing problem in instrumentation and measurement tasks involving cameras, LiDARs, and other spatial sensors. In circular-target-based LiDAR-camera extrinsic calibration, \revision{a }3D circular center measured from LiDAR and its corresponding 2D projected center measured in the image serve as cross-modal geometric observations for estimating the rigid transformation between the two sensor frames. \revision{Conventional pipelines can bias both measurements: }the 3D \revision{center is often obtained by sequential }plane fitting, point projection, and 2D circle fitting\revision{, while the image }ellipse center is frequently \revision{treated as the projected circle center even though the two generally differ under perspective projection}. This paper focuses on accurate 3D and 2D circular-center measurement and uses LiDAR-camera extrinsic calibration as a representative application. 
\revision{A conformal-geometric-algebra }estimator is integrated with RANSAC to jointly \revision{recover }the 3D center, normal, and radius from noisy or partially observed LiDAR points. \revision{A chord-length-variance criterion then estimates the }2D projected center, with \revision{its twofold }ambiguity resolved by homography validation or a quasi-RANSAC fallback. Synthetic and real-sensor experiments show that the proposed method improves circular-center measurement accuracy and reduces extrinsic calibration error across different target and sensor configurations.

\end{abstract}

\begin{IEEEkeywords}
Circular-center measurement, geometric measurement, LiDAR-camera extrinsic calibration, circular target, conformal geometric algebra.
\end{IEEEkeywords}

\section{Introduction}
\label{sec: intro}
\IEEEPARstart{A}{ccurate} measurement of circular centers is a recurring geometric sensing problem in instrumentation and measurement systems. Circular structures appear in calibration targets, holes, rings, wheels, reflectors, and manufactured parts, and their centers are often used as compact geometric observations for pose estimation, dimensional inspection, target localization, and multi-sensor registration. Although the circle is a simple geometric primitive, accurately measuring its center from real sensor data remains nontrivial when the observations are sparse, noisy, partially occluded, or affected by perspective projection.
\rev{The same primitive supports pose measurement for circular parts in robotic manipulation and assembly~\cite{wang2022pose}, while calibrated vision systems use geometric features to monitor robot pose and drift in precision digital manufacturing~\cite{hedstrand2024photogrammetry}. When a circular feature defines an equipment datum, its localization also limits the registration accuracy between the physical asset and its digital representation. In each case, center-measurement error propagates into the coordinate frame of the downstream task.}

This paper focuses on the measurement of 3D and 2D circular centers. The 3D problem arises when a spatial sensor, such as a LiDAR, observes boundary points of a circular object. A common solution is to fit a plane, project the points onto that plane, and then fit a 2D circle~\cite{Rusu_ICRA2011_PCL,Domhoficra,Beltran,tim_full_2025,liu_sensors_2023,Xingxing}. This decoupled procedure is convenient, but it does not directly estimate the spatial circle and may introduce center bias when the point cloud is sparse or only partially observed~\cite{Fremont,dorst}. The 2D problem arises when a camera observes a 3D circle under perspective projection. The projected contour is generally an ellipse, but the ellipse center is not, except under special viewing geometries, the projection of the physical 3D circle center~\cite{KimPAMI,cucci2016accurate,cctag}. Therefore, directly using the ellipse center as the measured 2D circular center introduces a systematic projective bias.

LiDAR-camera extrinsic calibration provides a representative and practically important application of this circular-center measurement problem. The measurand of extrinsic calibration is the rigid transformation between the LiDAR and camera coordinate frames. In circular-target-based calibration, this transformation is estimated from correspondences between the 3D circular centers measured by LiDAR and the corresponding 2D projected centers measured in the image. Any bias in either measurement channel propagates to the final extrinsic parameters and affects point-image registration, sensor fusion, and subsequent perception. Recent studies have addressed LiDAR-camera and heterogeneous-sensor calibration from different aspects, including keypoint-based online calibration~\cite{tim_ye_2022_rgkcnet}, adaptive voxelization for small-FoV LiDAR--camera systems~\cite{tim_liu_2022_adaptive_voxel}, overlap-free environmental calibration~\cite{tim_zhang_2023_overlapfree}, noise-aware calibration for nonrepetitive scanning LiDAR~\cite{tim_cui_2024_aclc}, scene-aware online calibration~\cite{tim_gong_2024_sceneaware}, vanishing-point-based calibration~\cite{tim_zhao_2024_vanishing}, transformer-based targetless calibration~\cite{tim_liu_2024_transformer}, and full-lifecycle calibration~\cite{tim_wei_2025_lifecycle}. These methods improve the calibration process from the viewpoints of feature extraction, scene utilization, sensor configuration, online correction, or lifecycle maintenance. In contrast, the present work focuses on a complementary but fundamental issue: the accurate measurement of circular-center primitives used to form 3D--2D calibration correspondences.

\begin{figure}[!t]
    \centering
    \includegraphics[width=1.0\linewidth]{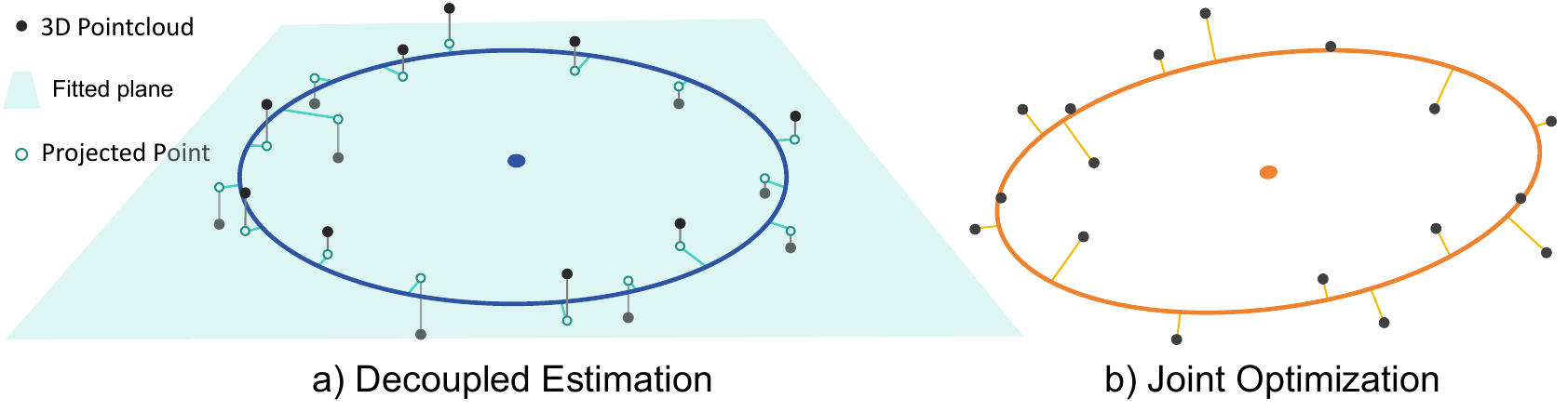}
    \caption{Comparison of 3D circle fitting approaches. Decoupled fitting~\cite{Rusu_ICRA2011_PCL} estimates a plane first, then fits a 2D circle on projected points. Joint optimization~\cite{Fremont,dorst} simultaneously minimizes point-to-plane and point-to-circle distances.}
    \label{fig:error-3d}
\end{figure}

\begin{figure}[!htbp]
    \centering
    \includegraphics[width=0.9\linewidth]{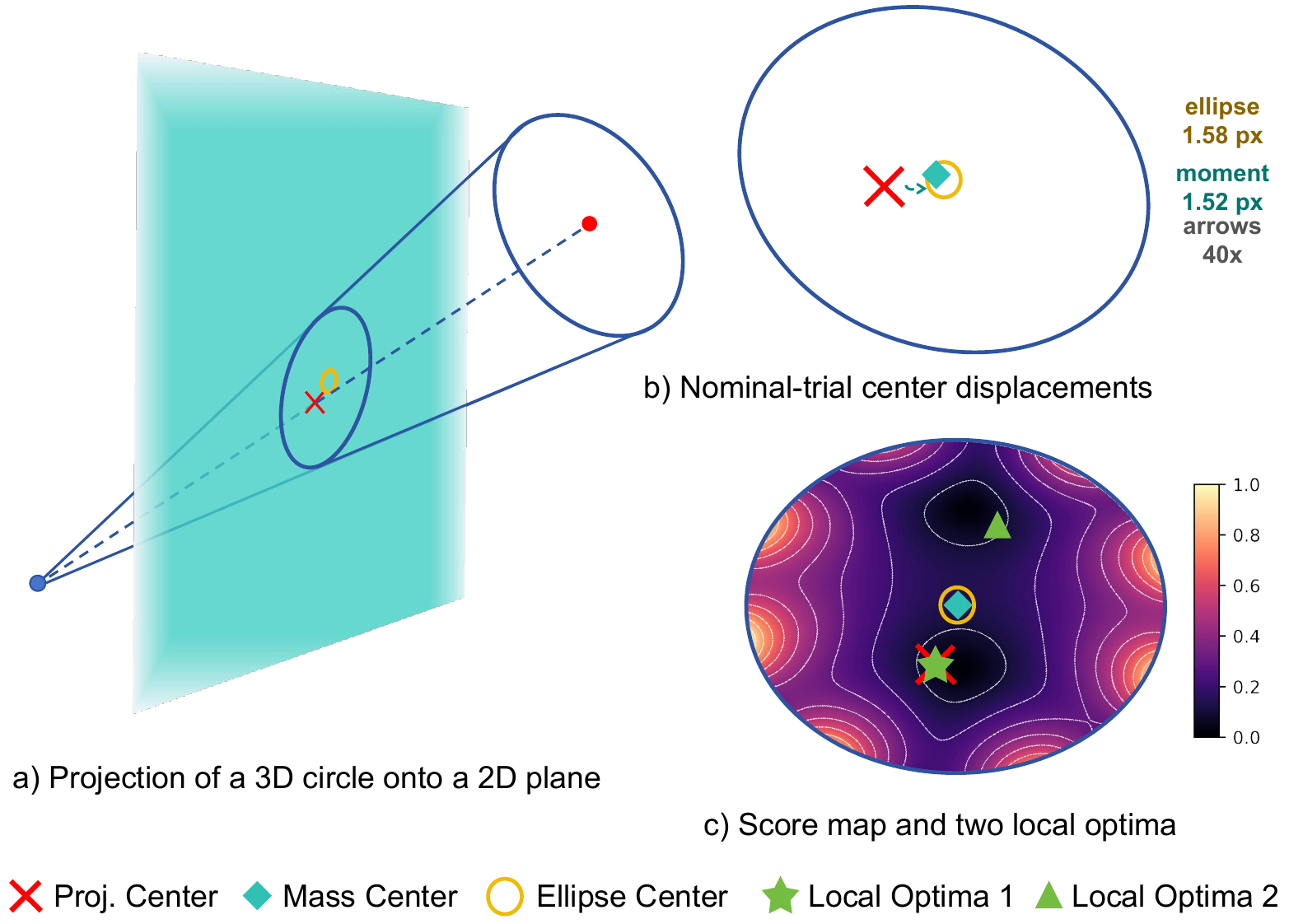}
    \caption{Locating the true projected center of a 3D circle from its 2D image. (a) A circle in 3D space projects to an ellipse. (b) \revision{In an example trial, the }projection of the physical 3D center (red cross) is displaced \revision{by 1.58~px }from the \revision{fitted-ellipse }center (yellow circle) and \revision{by 1.52~px from the }image-moment center (cyan diamond). \revision{the arrows are magnified $40\times$ for visibility. }(c) The proposed score-map optimization locates the true center (green star) and distinguishes it from a false candidate (green triangle).}
    \label{fig:error-2d}
\end{figure}

Among calibration targets, circular patterns are attractive because circular holes, printed disks, reflective rings, and spherical objects can be detected in both LiDAR and imagery under partial observation. \revision{Planar circular targets, in particular, are well suited for calibration as their geometry can be manufactured accurately and observed repeatedly.
}Their accuracy, however, depends on the correct measurement of a 3D circle center in LiDAR and its corresponding 2D projection in the image. This paper refers to this bottleneck as the \emph{Circle Center Problem}. As illustrated in~\cref{fig:error-3d,fig:error-2d}, it has two coupled sources of systematic error. First, many pipelines estimate a LiDAR circle center by fitting a plane, projecting points onto the plane, and then fitting a 2D circle~\cite{Rusu_ICRA2011_PCL,Domhoficra,Beltran,tim_full_2025,liu_sensors_2023,Xingxing}. Second, the 2D ellipse center is generally not the projection of the 3D circle center under perspective projection~\cite{KimRAL,A4LidarTag,tim_full_2025}. The resulting center bias degrades the \revision{3D-2D }correspondence set used in calibration and other circular-feature-based measurement tasks.


To address this problem, we propose two circular-center measurement methods and validate them through LiDAR-camera extrinsic calibration. First, conformal geometric algebra (CGA) is used to formulate 3D circle fitting as an eigenvalue problem in conformal space~\cite{dorst}. The resulting estimator jointly recovers the center, normal, and radius and can be embedded in RANSAC for robustness to noise, sparse samples, and outliers. Second, the true 2D projected center is estimated by minimizing the variance of unprojected chord lengths~\cite{KimPAMI,cucci2016accurate}. Because the loss may produce two local minima, we resolve the ambiguity by homography-based validation when coplanar circles are available, and by a quasi-RANSAC fallback otherwise. The contributions are summarized as follows:
\begin{itemize}
    \item We formulate circular-target LiDAR-camera calibration as a 3D-2D circular-center measurement problem and identify the two major bias sources in existing pipelines.
    \item We develop a CGA-RANSAC 3D circle estimator and a perspective-aware 2D projected-center estimator to improve correspondence accuracy.
    \item We validate the proposed method on numerical, synthetic, and real-world datasets, showing improved center measurement and extrinsic calibration accuracy over representative baselines. \revision{To facilitate reproducibility, the implementation is now publicly released}\footnote{\url{https://github.com/xiahaa/circular-center-calibration.git}}.
\end{itemize}
The rest of this paper is organized as follows. ~\cref{sec: related} reviews related work on LiDAR-camera calibration and circular-feature measurement. ~\cref{sec: method} details the proposed 3D and 2D circular-center measurement methods. ~\cref{sec:exp} presents the experimental setup and evaluation. Finally, ~\cref{sec: conclusion} concludes the paper and discusses future work.

\section{Related Work}
\label{sec: related}
\subsection{Target-Based and Targetless LiDAR-Camera Calibration}
Existing LiDAR-camera calibration methods can be broadly divided into target-based and targetless methods. Target-based methods use known objects such as checkerboards~\cite{geiger2012automatic}, spheres~\cite{Kummerle,ZhangRAL}, or fiducial/circular boards~\cite{tim_wei_2025_lifecycle,liu_sensors_2023,A4LidarTag,KimRAL,zheng2025fast,Domhoficra,Beltran,opencalib} to establish measurable geometric correspondences. These methods are accurate and repeatable, and are therefore widely used for sensor production, maintenance, and benchmark construction. However, their accuracy depends directly on how reliably the target primitives are measured in each sensing modality. For circular targets, this means that both the 3D circular center and its 2D projected counterpart must be measured accurately.
Targetless methods remove the target requirement by exploiting natural edges~\cite{yuan2021pixel,levinson2013automatic}, intensity correlations~\cite{Pandey}, motion constraints~\cite{TaylorTRO,tim_ye_2022_rgkcnet}, semantic or geometric features~\cite{SGCalib,tim_liu_2022_adaptive_voxel,tim_zhao_2024_vanishing,tim_gong_2024_sceneaware}, and learned depth or representation models~\cite{CMRNext,3DGScalib,tim_liu_2024_transformer}. 

\subsection{Circular Targets and Center Measurement}
Circular targets are frequently used because they are distinctive in images and remain observable in sparse LiDAR scans. Spherical targets provide view-invariant geometry and have been used for robust spatial center estimation~\cite{Kummerle,ZhangRAL}. Planar boards with circular holes or disks are also common because their boundaries can be detected in both modalities~\cite{Fremont,A4LidarTag,KimRAL,Beltran,Domhoficra,opencalib,zheng2025fast}. In many existing pipelines, the 3D circle center is obtained by a sequential process: plane fitting, point projection, and 2D circle fitting~\cite{Domhoficra,Beltran,tim_full_2025,liu_sensors_2023,Xingxing}. This procedure is convenient but does not directly solve the 3D circle fitting problem and may introduce center bias when the point cloud is noisy, sparse, or only partially observed~\cite{Fremont,dorst}. Some recent methods improve target extraction by template matching, edge dilation, or constrained optimization~\cite{opencalib,zheng2025fast,KimRAL}, but the center-measurement error remains a limiting factor. \revision{Recently, Circle-Net~}\cite{wei2022circlenet} \revision{proposes a deep method for detecting boundary points. However, their work focuses on end-to-end circle detection rather than center estimation. Moreover, to our knowledge, they have not released their model, making reproducibility difficult.
}The image-side center measurement is equally important. A 3D circle generally projects to an ellipse, but the ellipse center is not the projection of the 3D circle center except under special viewing geometries. Nevertheless, several methods still use the ellipse or blob center as the 2D correspondence~\cite{KimRAL,A4LidarTag,tim_full_2025}. Other methods project the center using poses estimated from auxiliary features such as ArUco markers or checkerboard corners~\cite{Beltran,Domhoficra,zheng2025fast,opencalib,liu_sensors_2023}, which transfers the accuracy requirement to another detection module. 
\revision{Cucci~}\cite{cucci2016accurate} \revision{refines the projected center of concentric circular targets through a geometric invariant, making it a directly relevant image-side baseline. Although they identify dual-minimum issues, they do not propose a solution. CCTag~}\cite{cctag} \revision{changes the fiducial design to concentric rings and optimizes the 2D projection center by maximizing rectified intensity contrast, thereby resolving projective ambiguity.
} This paper directly addresses both the LiDAR-side 3D center measurement and the image-side projected-center measurement within a unified calibration pipeline\revision{.
}

\subsection{\revision{Circular Features in Precision Engineering}}
\revision{Circular-feature pose measurement is used in calibration, manipulation of circular metal parts, robotic assembly, and localization~}\cite{wang2022pose}\revision{. In precision digital manufacturing, calibrated photogrammetric systems measure robot pose and monitor drift, so geometric feature error propagates directly into the manufacturing coordinate frame~}\cite{hedstrand2024photogrammetry}\revision{. This paper offers an accurate circular-center estimator that can support these measurement chains besides the representative Lidar-camera calibration task}.

\section{Method}
\label{sec: method}

\begin{figure*}[!htbp]
    \centering
    \includegraphics[width=0.9\linewidth]{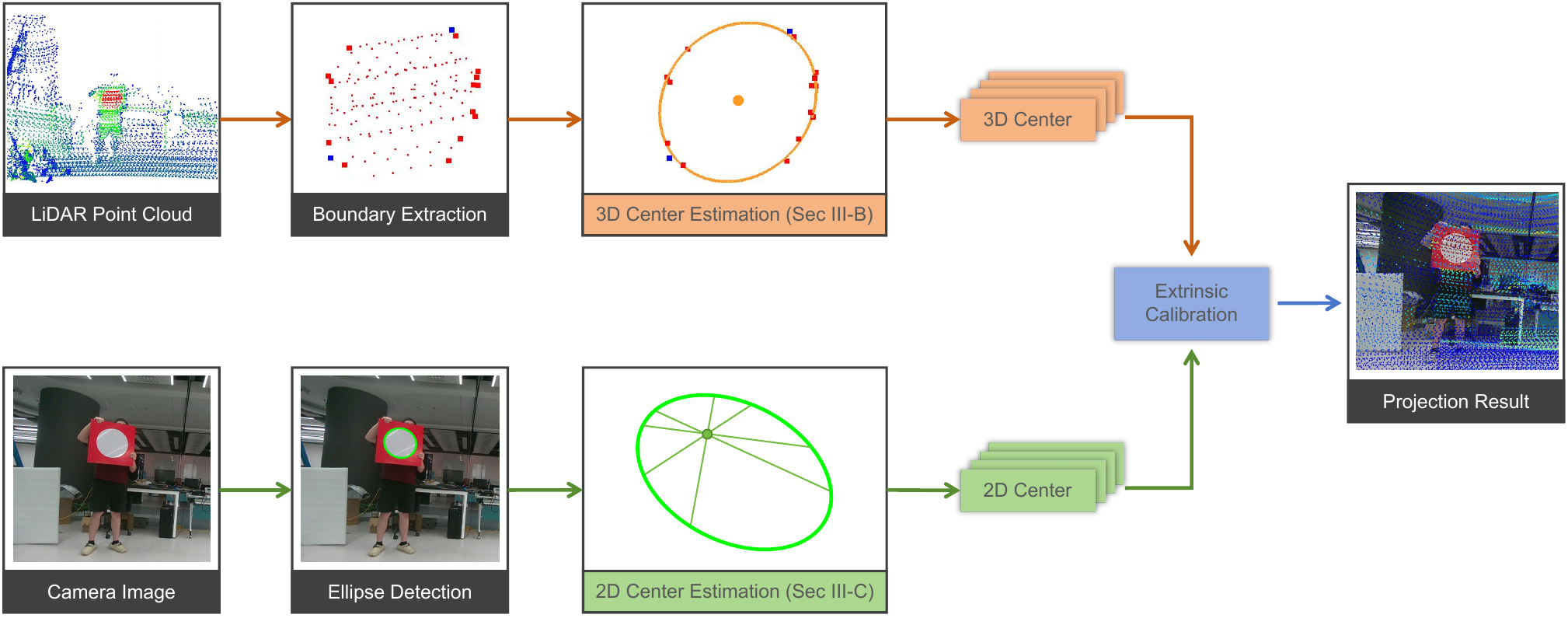}
    \caption{Overview of the proposed circular-center measurement methods and their application to LiDAR-camera extrinsic calibration. The main measured quantities are the 3D circular centers from LiDAR and the corresponding 2D projected centers in the image.}
    \label{fig:scheme}
\end{figure*}
\revision{Throughout }this paper, \revision{bold uppercase Roman letters denote matrices, bold }lowercase Roman letters \revision{denote vectors, and lowercase letters denote scalars. Capital italic letters denote conformal entities. The symbol }$\mathbf{0}$ \revision{denotes a zero matrix or vector of the required size, and }$\mathbf{I}$ \revision{denotes an identity matrix.
}

\subsection{Overview}
This section describes the proposed 3D and 2D circular-center measurement methods and their use in LiDAR-camera extrinsic calibration. We assume that the intrinsic parameters of the camera and LiDAR are known. The calibration application follows the pipeline in~\cref{fig:scheme}, where 3D circular centers measured from LiDAR and 2D projected centers measured from images are used to establish correspondences for extrinsic estimation. Since the primary contribution of this work lies in the measurement of these circular centers, we first briefly describe the feature preprocessing modules and then present the proposed 3D and 2D center estimators.

The LiDAR feature preprocessing module takes the point cloud captured by the LiDAR as its input. Its output is a list of boundary points that will be used for circle fitting. To obtain this output, distance and reflectance filtering are applied to segment the points of interest from the raw point cloud. Subsequently, distance clustering is employed to identify clusters of suitable size. Boundary points can then be extracted based on depth discontiuties for scanning LiDAR~\cite{Beltran} or directional gap among its neighbors~\cite{zheng2025fast}.

The image feature preprocessing module uses the corresponding image captured by the camera as its input. Due to the perspective effect, the projection of a 3D circle is typically an ellipse. Thus, the output of this module is a list of detected ellipses. For ellipse detection, we recommend using SOTA arc-based approaches \cite{elsd, aamed}. These approaches have demonstrated superior performance compared to hough transform based methods \cite{RHT,opencv_library}  \revision{in difficult contour-detection settings, especially under occlusion}, as shown in numerous studies \cite{elsd, aamed, lu2019arc}.

\revision{Given these correspondences, Perspective-$n$-Point }(PnP) \revision{estimates the six-degree-of-freedom }transformation $\mathbf{T}_{\mathrm{L}}^{\mathrm{C}}$ from the LiDAR frame to the camera frame. \revision{Let $\{\mathbf{p}_i^{\mathrm{L}}\}_{i=1}^n$ be the }3D \revision{centers described }in~\cref{sec: 3dc} and \revision{$\{\mathbf{q}_i^{\mathrm{C}}\}_{i=1}^n$ their image projections described in}~\cref{sec: 2dc}\revision{. The transformation is obtained }by minimizing the \revision{reprojection error:
}
\begin{equation}
    \mathbf{T}_{\mathrm{L}}^{\mathrm{C}}=\arg \min _{\mathbf{T}_{\mathrm{L}}^{\mathrm{C}}} \sum_{i=1}^n\left\|\mathbf{q}_i^{\mathrm{C}}-\pi(\mathbf{T}_{\mathrm{L}}^{\mathrm{C}} \mathbf{p}_i^{\mathrm{L}})\right\|_2,
    \label{eq1}
\end{equation}
where \revision{$\pi(\cdot)$ denotes perspective projection with the rectified camera intrinsics. }\rev{Because $\mathbf{p}_i^{\mathrm{L}}$ is measured in metric LiDAR coordinates, this 3D-2D formulation directly determines the translation scale. Lifting image observations into an additional set of 3D points is therefore unnecessary and would introduce depth-recovery error~\cite{scaramuzza2011vo}.}

\subsection{3D Center Estimation}
\label{sec: 3dc}
To robustly measure the 3D circle parameters, namely the center $\mathbf{c}$, normal $\mathbf{n}$, and radius $r$, from noisy boundary points ${p_i}\in \mathbb{R}^3,\ i=1,2,\cdots,n$, we propose a CGA-based estimator integrated within a RANSAC framework. Unlike classical decoupled approaches, i.e., plane fitting followed by 2D fitting, the proposed method jointly models the circle as a geometric primitive in conformal space, enabling direct minimization of geometric distance rather than algebraic proxies. This leads to significantly improved accuracy under noise and partial occlusion, as validated in~\cref{sec: exp_sub1}. Furthermore, by embedding this analytical solver into RANSAC~\cite{Fischler1981RandomSC}, our approach gains strong resilience to outliers without sacrificing efficiency.

CGA extends Euclidean space $\mathbb{R}^3$ into a 5D Minkowski space $\mathbb{R}^{4,1}$~\cite{HONGBO}, spanned by an orthonormal basis $\left\{\mathbf{e}_1, \mathbf{e}_2, \mathbf{e}_3, e_{+}, e_{-}\right\}$, where $e_{+}^2=1, e_{-}^2=-1$, and $e_{+} \cdot e_{-}=0$. For geometric clarity, it is often convenient to reparameterize this space using the null basis:
\begin{equation}
n_0=\frac{1}{2}\left(e_{-}+e_{+}\right), \quad n_{\infty}=e_{-}-e_{+},
\end{equation}
satisfying $n_0 \cdot n_0=n_{\infty} \cdot n_{\infty}=0$ and $n_0 \cdot n_{\infty}=-1$. A 3D point $\mathbf{p}=(x, y, z)$ is mapped to its conformal representation:
\begin{equation}
P=\mathbf{p}+n_0+\frac{1}{2}\|\mathbf{p}\|^2 n_{\infty}
\end{equation}
where $n_0$ encodes the origin and $n_{\infty}$ represents the point at infinity. Crucially, the squared Euclidean distance between two points $\mathbf{p}, \mathbf{q}$ is recovered via the inner product of their conformal counterparts:
\begin{equation}
P \cdot Q=-\frac{1}{2}\|\mathbf{p}-\mathbf{q}\|^2 \ .
\end{equation}
This \revision{representation converts Euclidean distances into bilinear conformal products}. In CGA, a 3D circle $CC$ is represented as the outer product ($\wedge$) of a sphere $SP$ and a plane $PL$:
\begin{equation}
\label{eq:cga_objects} 
\begin{aligned}
SP &= C - 0.5\rho^2 n_{\infty},\  
PL = \mathbf{n} + \delta n_{\infty}, \\
CC &= SP \wedge PL.
\end{aligned}
\end{equation}
Here, $SP$ represents a sphere centered at $\mathbf{c}$ with radius $\rho$\revision{, and }$PL$ represents a plane with normal $\mathbf{n}$ and offset $\delta$. A point $P$ lies on $CC$ if and only if it satisfies both \revision{$P\cdot SP=0$ and $P\cdot PL=0$. The CGA metric yields the following scalar residual}~\cite{dorst}:
\begin{equation}
-\frac{(P \cdot CC) *(P \cdot CC)}{CC * CC}=\frac{(P \cdot SP)^2}{SP^2}+\frac{(P \cdot PL)^2}{PL^2} .
\label{eq:cga-distance}
\end{equation}
where $*$ denotes the scalar product. This measure in~\eqref{eq:cga-distance} provides a geometrically meaningful cost function: the first term approximates the deviation from the spherical constraint, while the second enforces exact planar alignment. Unlike algebraic errors used in many direct estimation methods, this cost reflects the true spatial relationship between the point and the circle. Therefore, to fit a circle to $n$ noisy 3D points $\left\{\mathbf{p}_i\right\}$, we minimize the sum of approximate squared distances from points to the circle, which can be writen as
\begin{equation}
\mathcal{L}(CC)=\frac{1}{n} \sum_{i=1}^n \frac{\left(P_i \cdot CC\right) *\left(P_i \cdot CC\right)}{CC * CC} \ .
\end{equation}
\revision{The minimizer can be recovered analytically from a fixed-size eigensystem. }\rev{Before forming the conformal coordinates, we translate the Euclidean samples to their centroid and normalize them by their root-mean-square distance. The fitted circle is then mapped back to the input frame. This normalization improves conditioning under large coordinate offsets or changes of scale without imposing a regularization prior.} \revision{Define }the data matrix as:
\begin{equation}
\mathbf{D}=\left[\begin{array}{ccc}
\mathbf{p}_1 & \cdots & \mathbf{p}_N \\
1 & \cdots & 1 \\
\frac{1}{2}\left\|\mathbf{p}_1\right\|^2 & \cdots & \frac{1}{2}\left\|\mathbf{p}_n\right\|^2
\end{array}\right],
\label{eq: D}
\end{equation}
and the metric tensor: 
\begin{equation}
\mathbf{M}=\left[\begin{array}{ccc}
\mathbf{I}_3 & \mathbf{0} & \mathbf{0} \\
\mathbf{0}^{\top} & 0 & -1 \\
\mathbf{0}^{\top} & -1 & 0
\end{array}\right] \ \revision{. 
}\label{eq: M}
\end{equation}
\rev{Let $\mathbf{S}=\frac{1}{n}\mathbf{D}\mathbf{D}^{\top}$ and $\mathbf{P}=\mathbf{S}\mathbf{M}$. Although $\mathbf{S}$ is symmetric positive semidefinite, $\mathbf{P}$ is generally not symmetric under the Euclidean inner product. It instead satisfies $\mathbf{P}^{\top}\mathbf{M}=\mathbf{M}\mathbf{P}$ and is self-adjoint with respect to the indefinite CGA metric. The desired circle is encoded by the real two-dimensional invariant subspace associated with the spectrum nearest zero in magnitude. In implementation, we recover this near-zero subspace, check the rank and spectral separation, and reject a sample if a real two-dimensional subspace cannot be identified reliably.}
\revision{Let $V,U\in\mathbb{R}^5$ be }\rev{a basis of this real near-zero invariant subspace}. Their outer product yields the dual bivector representation of the circle:
\begin{equation}
    E=V \wedge U=\left[\begin{array}{ccc}
{\left[\mathbf{v}^{\times}\right]} & \mathbf{0} & \mathbf{0} \\
v_{0}\mathbf{I}_3 & -\mathbf{v} & 0 \\
-v_{\infty}\mathbf{I}_3 & 0 & \mathbf{v} \\
\mathbf{0}^T & -v_{\infty} & v_0
\end{array}\right]\left[\begin{array}{c}
\mathbf{u} \\
u_0 \\
u_{\infty}
\end{array}\right] .
\end{equation}
The circle parameters are then extracted from $E$ using closed-form formulas:
\begin{equation}
\begin{aligned}
\mathbf{n} & =-E_{3,4,5}, \quad \mathbf{c} =(\mathbf{K n}) /\|\mathbf{n}\|^2 \ , \\
r^2 & =\|\mathbf{c}\|^2-2 \mathbf{n} \cdot E_{6,7,8}-2(\mathbf{c} \cdot \mathbf{n})^2 \ ,
\end{aligned}
\end{equation}
where $E_i$ denotes the $i$-th component of $E$ in the ordered basis $\left\{\mathbf{e}_1, \mathbf{e}_2, \mathbf{e}_3, n_0, n_{\infty}\right\}$, and $\mathbf{K}$ is the $3 \times 3$ matrix defined as $\mathbf{K}=-E_9\mathbf{I}+[E_1,E_2,E_3]^\top_\times$.

\revision{The fit requires one fixed-size eigendecomposition }and no iterative optimization, yielding a fast and deterministic fit. For a rigorous derivation, we refer the reader to read~\cite{dorst}. \rev{It is important to observe that the basis of the two-dimensional subspace is not unique, i.e. swapping $V$ and $U$, changing their signs, or rescaling them by nonzero factors changes the bivector representation but not the represented circle. In the extraction formulas, the corresponding signs of $\mathbf{n}$ and $\mathbf{K}$ cancel in $\mathbf{c}$, while the radius is unchanged because its sign-dependent terms are paired or squared. Only the orientation of the normal remains ambiguous. We therefore evaluate it as an unoriented direction and choose a deterministic representation by making its largest-magnitude component positive.}
It is inevitable that real-world point clouds contain outliers arising from sensor noise, occlusions, or spurious detections. While our CGA estimator excels in noisy but inlier-dominated scenarios, robustness to gross outliers is essential for deployment in unstructured environments. Fortunately, \revision{as }our solution reduces circle fitting to a linear eigenvalue problem, similar in structure to the eight-point~\cite{Hartley} and five-point~\cite{Nister} algorithms for essential matrix estimation, it integrates seamlessly with RANSAC~\cite{Fischler1981RandomSC}. To construct our RANSAC framework, we leverage three key components:
\begin{enumerate}
\item \textbf{Minimal sample}: \revision{five points are used to generate a circle hypothesis through}~\eqref{eq: D} and~\eqref{eq: M}\revision{.
}\item \textbf{Model \revision{estimation}}: \revision{each hypothesis is obtained by the fixed-size eigendecomposition.
}\item \textbf{Inlier scoring}: \revision{the residual in~\eqref{eq:cga-distance} measures agreement with both the spherical and planar constraints}.
\end{enumerate}
\rev{For $n$ input points, accumulating the fixed $5\times5$ scatter matrix costs $O(n)$ time and $O(1)$ auxiliary memory; the $5\times5$ eigensolve has constant cost. A RANSAC fit with $I$ hypotheses therefore costs $O(In)$ in the worst case. Our implementation rejects coincident or rank-deficient minimal samples, refits the consensus set, and updates $I$ from the observed inlier ratio and requested confidence.} This design inherits the speed and stability of CGA while gaining outlier rejection capability.

\begin{figure}
    \centering
    \includegraphics[width=0.5\linewidth]{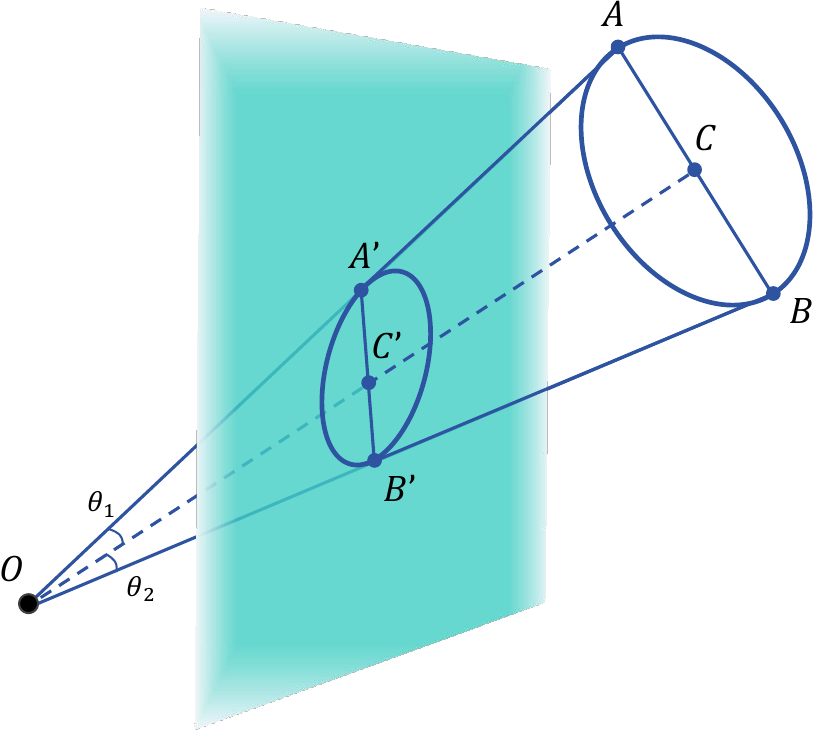}
    \caption{Geometric constrint under projection}
    \label{fig:circle_geo}
\end{figure}

\subsection{2D Center Estimation}
\label{sec: 2dc}
\rev{In the following, we assume the 2D estimator operate in a rectified image space by either rectifying the image or those contour points as lens distortion generally destroys the conic property~\cite{song2024distorted}.} \revision{Assuming }we are given a set of detected ellipses in the image, each parameterized by a conic matrix $\mathbf{Q} \in \mathbb{R}^{3 \times 3}$ such that a point $\mathbf{p}=[u, v, 1]^{\top}$ lies on the ellipse if $\mathbf{p}^{\top} \mathbf{Q} \mathbf{p}=0$, where $\mathbf{Q}=\left[Q_{11},\ Q_{12},\ Q_{13};Q_{21},\ Q_{22},\ Q_{23};Q_{31},\ Q_{32},\ Q_{33}\right]$ and 
$Q_{ij}$ denotes \revision{a conic coefficient. The }geometric center of \revision{the image ellipse is often used as the projected }3D circle \revision{center~}\cite{KimRAL,A4LidarTag,tim_full_2025}\revision{, but the two generally differ under }perspective projection~\cite{KimPAMI,cucci2016accurate,cctag}. We therefore seek the projection of the physical circle center, denoted by \revision{$\mathbf{c}\in\mathbb{R}^2$}.

Inspired by~\cite{cucci2016accurate}, we exploit the following key insight: if $\mathbf{c}$ is the true projection of a 3D circle center, then every chord passing through $\mathbf{c}$ in the image must correspond to a diameter in 3D space. Consequently, the distance from the camera center to the circular plane, calculated using trigonometry from the chord and $\mathbf{c}$ using trigonometry should exhibit zero variance. To quantify this, a loss function can be defined based on the variation of the recovered distances, computed over multiple candidate directions. As shown in~\cref{fig:circle_geo}, given a candidate center $c$ and an ellipse $\mathbf{Q}$, we sample $n$ lines passing through $c$ at uniformly spaced angles $\gamma_i$. Each line intersects the ellipse at two points $a_i, b_i$, whose deprojected rays in 3D (via inverse camera projection $\mathbf{r}_u=\mathbf{K}^{-1} \tilde{\mathbf{u}} /\left\|\mathbf{K}^{-1} \tilde{\mathbf{u}}\right\|$ where $\tilde{\mathbf{u}}$ is the homogeneous coordinate of $\mb{u}$) form a chord of the unknown 3D circle. Using the law of cosines on the viewing rays and the known circle radius $r$, we compute the distance $d_i=\|OC\|$ from the camera center $O$ to the 3D circle center $C$ ($s(\cdot)$ and $c(\cdot)$ are used to denote $\sin$ and $\cos$):
\begin{equation}
\begin{aligned}
d_i=\frac{\sqrt{2} r s \left(\theta_{i 1}+\theta_{i 2}\right)}{\sqrt{3-2 c \left(2 \theta_{i 1}\right)-2 c \left(2 \theta_{i 2}\right)+c \left(2\left(\theta_{i 1}+\theta_{i 2}\right)\right)}},\end{aligned}
\label{eq: oc}
\end{equation}
where $\theta_{i 1}=\cos ^{-1}\left(\mathbf{r}_{a_i} \cdot \mathbf{r}_{c}\right)$ and $\theta_{i 2}=\cos ^{-1}\left(\mathbf{r}_{b_i} \cdot \mathbf{r}_{c}\right)$ are the angular separations between the viewing rays and the candidate center direction. If $c$ is correct, all $d_i$ must be equal, which yields minimal variance. We thus minimize:
\begin{equation}
\mathcal{L}\left(c\right)=\frac{1}{n} \sum_{i=1}^n\left(d_i-\mu_d\right)^2, \quad \mu_d=\frac{1}{n} \sum_{i=1}^n d_i.
\label{eq: loss2d}
\end{equation}
Namely, for a given ellipse, we search for $\mb{c}$ for which the dispersion of the values of $d$ is minimized. Unlike~\cite{cucci2016accurate}, we avoid using the standard deviation to preserve smoother differentiability. However, this loss function typically exhibits two local minima within the ellipse with only one corresponding to the true center. Without additional constraints, these minima are indistinguishable based on the loss alone, a limitation not resolved in~\cite{cucci2016accurate}. In the following, we show that we could resolve this ambiguity using a second co-planar, non-concentric ellipse. The key observation is that a correct center enables a \textit{canonical rectifying homography} $\mathbf{H}$ that transforms the ellipse into a circle centered at the origin, recovering the Euclidean geometry of the plane up to similarity (rotation and scaling)~\cite{cctag}. 
Crucially, if the candidate center $c$ is incorrect, for instance, when using the ellipse's geometric center as the projection center, the homography $\mathbf{H}$ still maps the associated ellipse to a circle, but it distorts the Euclidean geometry of the plane. As a result, while individual ellipses are rectified to circles, the relative scale between two co-planar, non-concentric ellipses is no longer preserved under $\mathbf{H}$, making it possible to distinguish the true center.

Specifically, given a candidate center $c^{\prime}$, we first apply an affine transformation $\mathbf{T}=\mathbf{A}_1 \mathbf{A}_2$ to center and rotate the ellipse into canonical form:
\begin{equation}
\mathbf{A}_1 = \left[\begin{array}{ccc}
\cos \theta & \sin \theta & 0 \\
-\sin \theta & \cos \theta & 0 \\
0 & 0 & 1
\end{array}\right], 
\mathbf{A}_2 = \left[\begin{array}{ccc}
1 & 0 & -e_x \\
0 & 1 & -e_y \\
0 & 0 & 1
\end{array}\right]
\end{equation}
where $\left(e_x, e_y\right)$ is the ellipse center and $\theta$ \revision{is its }orientation. The transformed conic becomes \revision{diagonal}:
\begin{equation}
\mathbf{Q}^\prime=\left(\mathbf{T}^{-1}\right)^{\top} \mathbf{Q T}^{-1}=\left[\begin{array}{ccc}
Q_{11} & 0 & 0 \\
0 & Q_{22} & 0 \\
0 & 0 & -1
\end{array}\right]
\end{equation}
The rectifying homography $\mathbf{H}=\mathbf{H}_e \mathbf{H}_a \mathbf{H}_p$ is then constructed as:
\begin{equation}
\begin{aligned}
    \mathbf{H}_p &= \begin{bmatrix}
1 & 0 & 0 \\
0 & 1 & 0 \\
Q^\prime_{11}c_x^{\prime} & Q^\prime_{22}c_y^\prime & Q^\prime_{33}
\end{bmatrix}\ , \\
\mathbf{H}_a &= \begin{bmatrix}
1/b & -a/b & 0 \\
0 & 1 & 0 \\
0 & 0 & 1
\end{bmatrix}, \ 
\mathbf{H}_e = \begin{bmatrix}
1 & 0 & x \\
0 & 1 & y \\
0 & 0 & 1
\end{bmatrix} \ , 
\end{aligned}
\end{equation}
with parameters $a, b, x, y$ derived from $Q^\prime_{i j}$ and the transformed center $\left[c_x^{\prime}, c_y^{\prime},1\right]^\top=\mathbf{T}\left[c_x, c_y, 1\right]^{\top}$ as follows:
\begin{equation}
\begin{aligned}
a &= \frac{Q^\prime_{22}c^{\prime}_x c^{\prime}_y}{Q^\prime_{11}{c^{\prime}_x}^2 + Q^\prime_{33}}\ , \\
b &= \sqrt{\frac{Q^\prime_{22}Q^\prime_{33}}{Q^\prime_{11}} \cdot \frac{Q^\prime_{11}{c^{\prime}_x}^2 + Q^\prime_{22}{c^{\prime}_y}^2 + Q^\prime_{33}}{(Q^\prime_{11}{c^{\prime}_x}^2 + Q^\prime_{33})^2} - a^2}\ , \\
x &= \frac{-c^{\prime}_x/b + c^{\prime}_y a/b}{Q^\prime_{11}{c^{\prime}_x}^2 + Q^\prime_{22}{c^{\prime}_y}^2 + Q^\prime_{33}} \ , \\
y &= \frac{-c^{\prime}_y}{Q^\prime_{11}{c^{\prime}_x}^2 + Q^\prime_{22}{c^{\prime}_y}^2 + Q^\prime_{33}} \ .
\end{aligned}
\end{equation}

Applying $\mathbf{H}$ to coplanar ellipses maps them to the rectified plane. If \revision{$\mathbf{c}$ is the physical }projection center, the ratio of their radii is preserved under similarity. \revision{We therefore select the candidate whose rectified radius ratio is closest }to the known physical \revision{ratio $\hat{r}_1/\hat{r}_2$. Its effectiveness is evaluated }in~\cref{sec:exp}.

\subsection{\revision{Quasi-RANSAC Without }Coplanar Circles}
In practical scenarios, coplanar circles may not always be available, as observed in our experiments (Section~\ref{sec:exp}). To enhance the applicability, we now propose a robust fallback strategy that enables calibration even when no such geometric constraint exists. For each detected ellipse, we optimize the 2D center loss function \eqref{eq: loss2d} using \rev{a three-level coarse-to-fine search} to identify both candidate centers corresponding to the two local minima. This yields 2n potential 2D centers. Since no additional geometric constraint is available to distinguish the true from spurious candidates, we integrate these ambiguities directly into a modified RANSAC framework for PnP. Unlike standard RANSAC, which randomly samples point correspondences, our approach samples candidate pairs: for each selected 3D point, we randomly choose one of its two 2D hypotheses. We iteratively construct pose hypotheses by sampling \rev{$s=4$} such ambiguous correspondences and evaluate the re-projection error. The hypothesis with the lowest cumulative error is retained. This procedure effectively explores the combinatorial space of candidate correspondences in practice.
\rev{Let $q$ be the fraction of valid 3D-2D correspondences and assume initially that either candidate is correct with probability $1/2$. A minimal PnP sample is entirely correct with probability $p=(q/2)^s$. Hence the number of independent trials required for confidence $\eta$ is
\begin{equation}
N=\left\lceil\frac{\log(1-\eta)}{\log\left(1-(q/2)^s\right)}\right\rceil.
\label{eq:quasi_ransac_iterations}
\end{equation}
We update $q$ from the best consensus and cap the search at a certain number of iterations.} As demonstrated in Section~\ref{sec:exp}, this approach significantly improves success rate over naive selection strategies, enabling reliable calibration in unstructured environments where coplanar features are absent.

\section{Experiments}
\label{sec:exp}

The experiments are designed to evaluate the proposed methods at two levels. First, we directly assess the accuracy of the measured 3D and 2D circular centers. Second, we use \revision{LiDAR-camera }extrinsic calibration as a representative downstream application to verify how circular-center measurement accuracy affects estimated extrinsic parameters and point-image alignment.

\subsection{\revision{Direct Circular-Center Measurement}}
\subsubsection{3D Circle Center Measurement}
\label{sec: exp_sub1}
We first evaluate the CGA-RANSAC 3D circle estimator against the PCL circle fitting implementation~\cite{Rusu_ICRA2011_PCL}, which follows a decoupled plane-fitting and 2D circle-fitting strategy. For each Monte Carlo trial, a random circle is generated with $\mathbf{c}_{\mathrm{gt}}\sim\mathcal{U}(-2,2)^3$, $r_{\mathrm{gt}}\sim\mathcal{U}(1,5)$ \revision{on a rotated plane with the normal }$\mathbf{n}_{\mathrm{gt}}\sim\mathcal{N}(0,\mathbf{I}_3)$. Points on the circle are corrupted by isotropic Gaussian noise. Four configurations are tested: full-circle sampling with isotropic noise, partial-arc sampling, sparse clustered sampling, and symmetric sparse-arc sampling. The center error is computed as $E_c=\|\mathbf{c}_{\mathrm{est}}-\mathbf{c}_{\mathrm{gt}}\|_2$ over 1000 trials.

\begin{figure}[!t]
    \centering
    \includegraphics[width=1\linewidth]{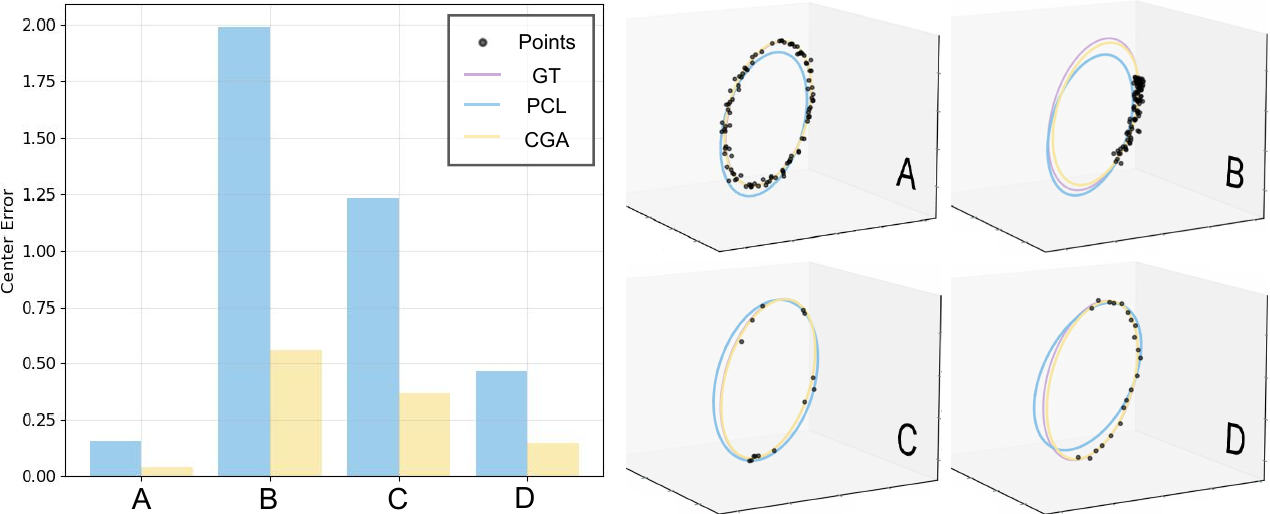}
    \caption{3D center measurement error under noise, partial coverage, and sparse sampling.}
    \label{fig:3d_exp1}
\end{figure}

Outlier robustness is evaluated by adding random outliers to a noisy full circle, with the outlier ratio varying from $10\%$ to $50\%$. Both methods use identical input point clouds and the same RANSAC iteration budget. As shown in~\cref{fig:3d_exp1,tb:exp_3d_outlier}, the proposed method consistently yields lower center error, especially under partial coverage, sparse sampling, and outliers. This confirms that directly estimating the 3D circle in conformal space is more stable than a decoupled fitting pipeline.

\begin{table}[!t]
\caption{Mean \revision{3D }Center Error Under Different Outlier Ratios}
\centering
\begin{tabular}{c|ccccc}
\midrule
\textbf{Method} & \multicolumn{5}{c}{\textbf{Outlier Ratio}} \\ \midrule
 & 10\% & 20\% & 30\% & 40\% & 50\% \\ \midrule
PCL & 0.0817 & 0.0870 & 0.0826 & 0.0775 & 0.0777 \\
CGA & \cellcolor{lightgreen}0.0354 & \cellcolor{lightgreen}0.0347 & \cellcolor{lightgreen}0.0356 & \cellcolor{lightgreen}0.0362 & \cellcolor{lightgreen}0.0364 \\ \midrule
\end{tabular}
\label{tb:exp_3d_outlier}
\end{table}

\rev{
The preceding trials vary one observation property at a time. To characterize their interaction, a second experiment uses a physical radius of $0.12$~m and jointly varies point count ($5$--$128$), visible arc ($45^\circ$--$360^\circ$), noise, and sampling distribution. Four distributions isolate the effect of angular support. \emph{Full-circle uniform} draws points over the entire circumference and serves as a coverage reference; consequently, the visible-arc variable is inactive for this case. \emph{Single arc} draws all points uniformly from one contiguous arc, whereas \emph{dual opposite arcs} splits the same total angular coverage between two arcs whose centers differ by $180^\circ$. Finally, the \emph{HDL-64-type bands} case quantizes the available arc into $B=\min(16,\max(2,\lfloor\sqrt{N}\rfloor))$ angular bands and adds small within-band jitter. This last distribution emulates the clustered circle intersections produced by a fixed multi-beam scanning pattern.~\cref{fig:3d_stress} shows that angular diversity is more important than point count alone. Both full-circle and dual-opposite-arc sampling are nearly saturated even with sparse points. For a single continuous arc, CGA reaches a success probability of 0.97 with eight points over $90^\circ$ and reaches 1.00 with 16 points over the same arc, whereas PCL requires at least 32 points and approximately $120^\circ$ of support to reach 0.99. The banded case is more demanding: CGA becomes reliable from 16 points and $60^\circ$, while PCL requires about 32 points and $120^\circ$. Increasing the number of points within only $45^\circ$--$60^\circ$ does not recover PCL performance because the added samples remain concentrated in the same angular bands. Thus, although both methods benefit from additional samples, CGA shows better resilience to limited angular coverage.

\begin{figure*}[!t]
    \centering
    \includegraphics[width=0.98\textwidth]{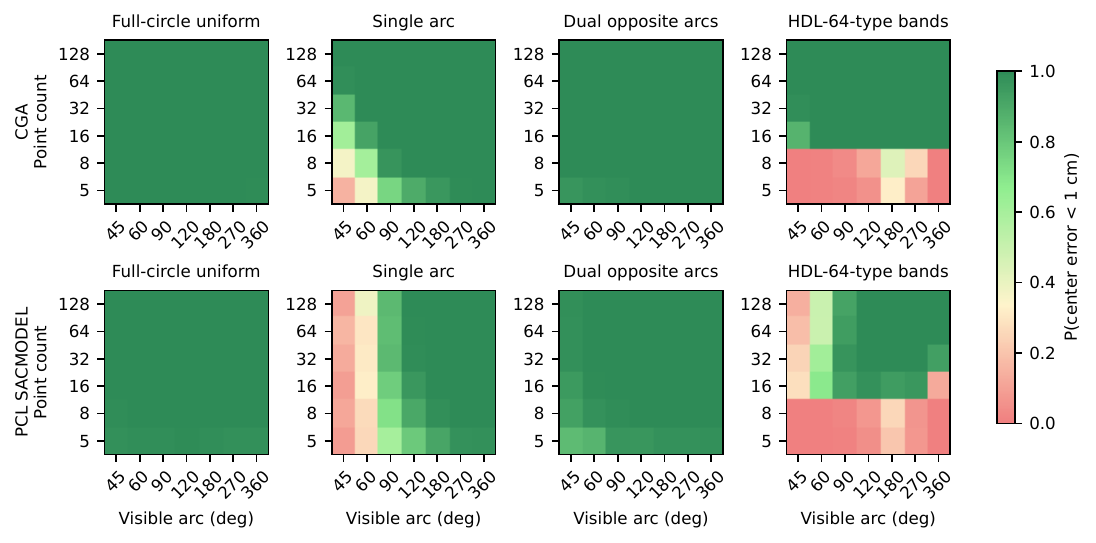}
    \caption{Probability of 3D center error below 1~cm as point count, visible arc, and sampling distribution vary at $\sigma=0.005r$. The visible-arc variable is inactive for the full-circle reference. Each cell contains 300 trials.}
    \label{fig:3d_stress}
\end{figure*}

To quantify tolerance to non-planar and non-circular targets, we additionally perturb the target away from an ideal planar circle. Normal-direction warp amplitudes and in-plane ellipse-axis deviations are swept from zero to $0.05r$. \cref{fig:target_tolerance} quantifies the resulting center bias and thereby relates target quality to attainable measurement accuracy. As can be seen, the effects of normal warp are small. At $0.05r$, the mean error remains below 1~cm for all three methods, with normalized CGA at approximately 5.5~mm. By contrast, non-circularity is more influential. At an axis deviation of 0.05, normalized CGA remains just below 1~cm, whereas CGA-RANSAC and PCL reach approximately 14.0 and 14.7~mm, respectively. These results also give practical engineering guidance for fabricating the calibration board, outlining the needed flatness and circularity requirements.

\begin{figure}[!t]
    \centering
    \includegraphics[width=\linewidth]{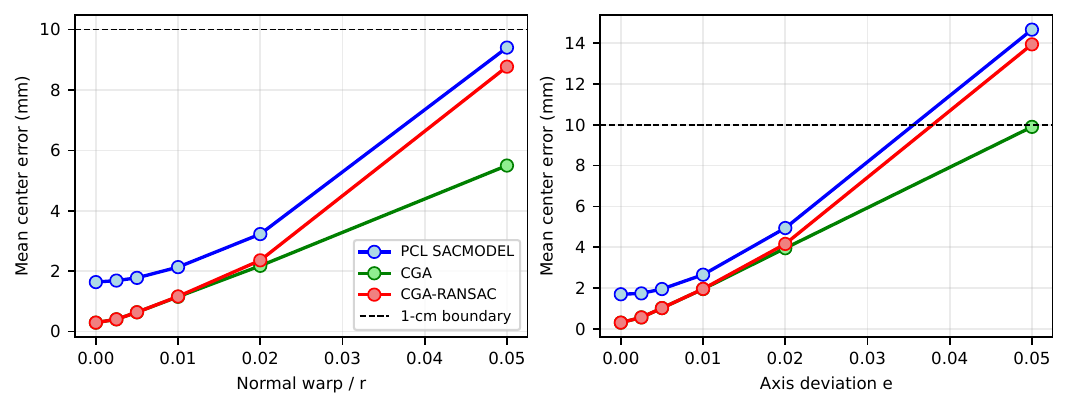}
    \caption{Sensitivity of 3D center measurement to nonplanar warp and noncircular axis error. The dashed line marks the 1-cm boundary.}
    \label{fig:target_tolerance}
\end{figure}
}

\subsubsection{2D Projected-Center Measurement}
\label{sec:2d_validation}
We next evaluate the 2D projected-center estimator using 1000 Monte Carlo trials. A synthetic camera with $f_x=f_y=600$, $c_x=640$, and $c_y=480$ observes two coplanar nonconcentric circles, where one circle is used for center estimation and the other for homography validation. Projected points are perturbed by Gaussian noise with $\sigma=1$ pixel, and ellipses are fitted by least squares. We compare the ellipse center, the center of mass, and the proposed refined center. The ground truth is the exact projection of the 3D circular center from the noise-free geometry.

\begin{figure}[!t]
    \centering
    \includegraphics[width=\linewidth]{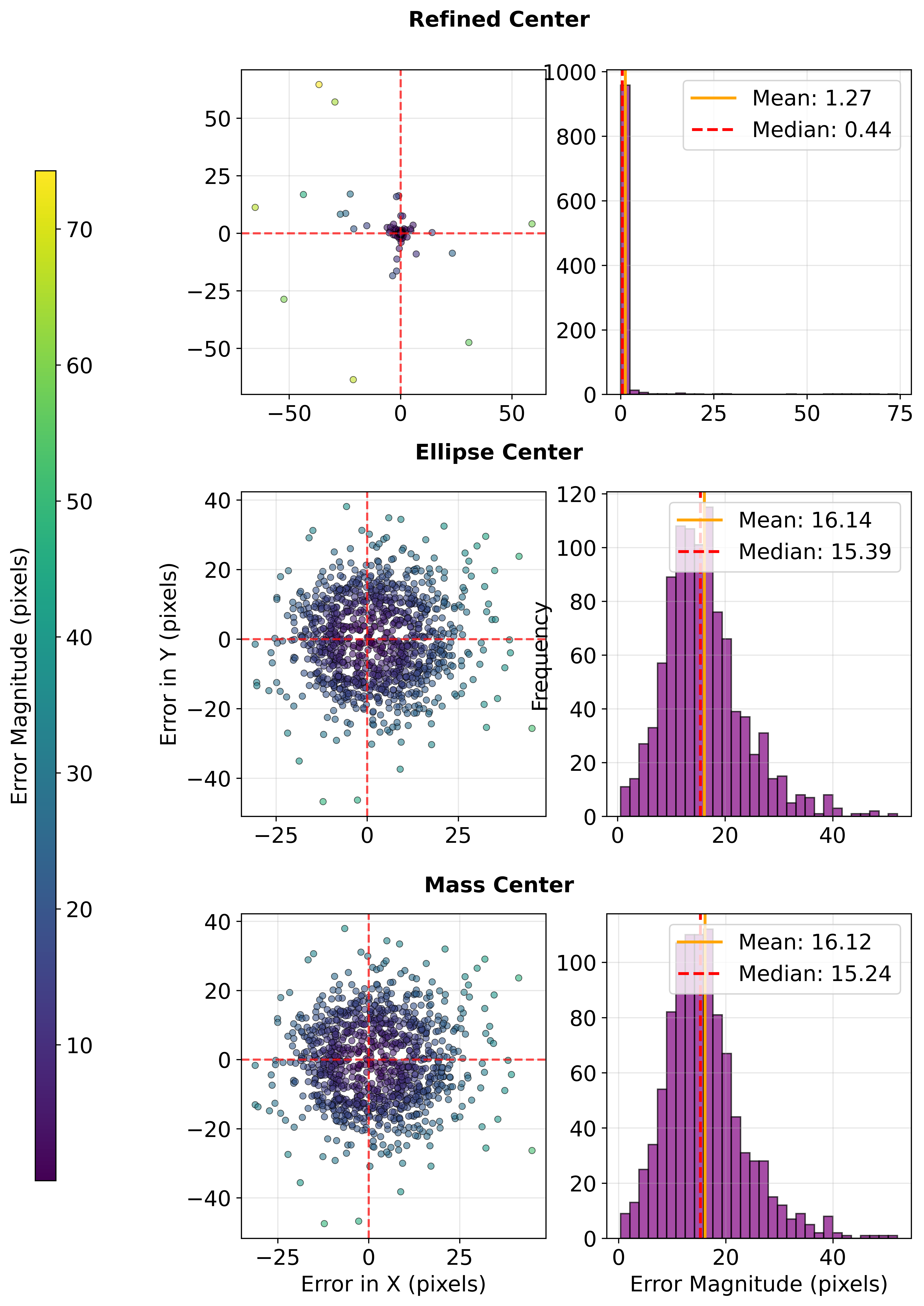}
    \caption{Distribution of 2D projected-center measurement errors.}
    \label{fig:2d_error}
\end{figure}
As shown in~\cref{fig:2d_error}, the ellipse \revision{and moment centers retain the projective displacement described in~}\cref{fig:error-2d}, whereas the proposed refinement \revision{yields }a lower mean error and a tighter distribution. \revision{Its remaining large errors arise mainly when image noise changes the ordering of the two }minima in the \revision{score }map.

\begin{figure}[!t]
    \centering
    \includegraphics[width=1\linewidth]{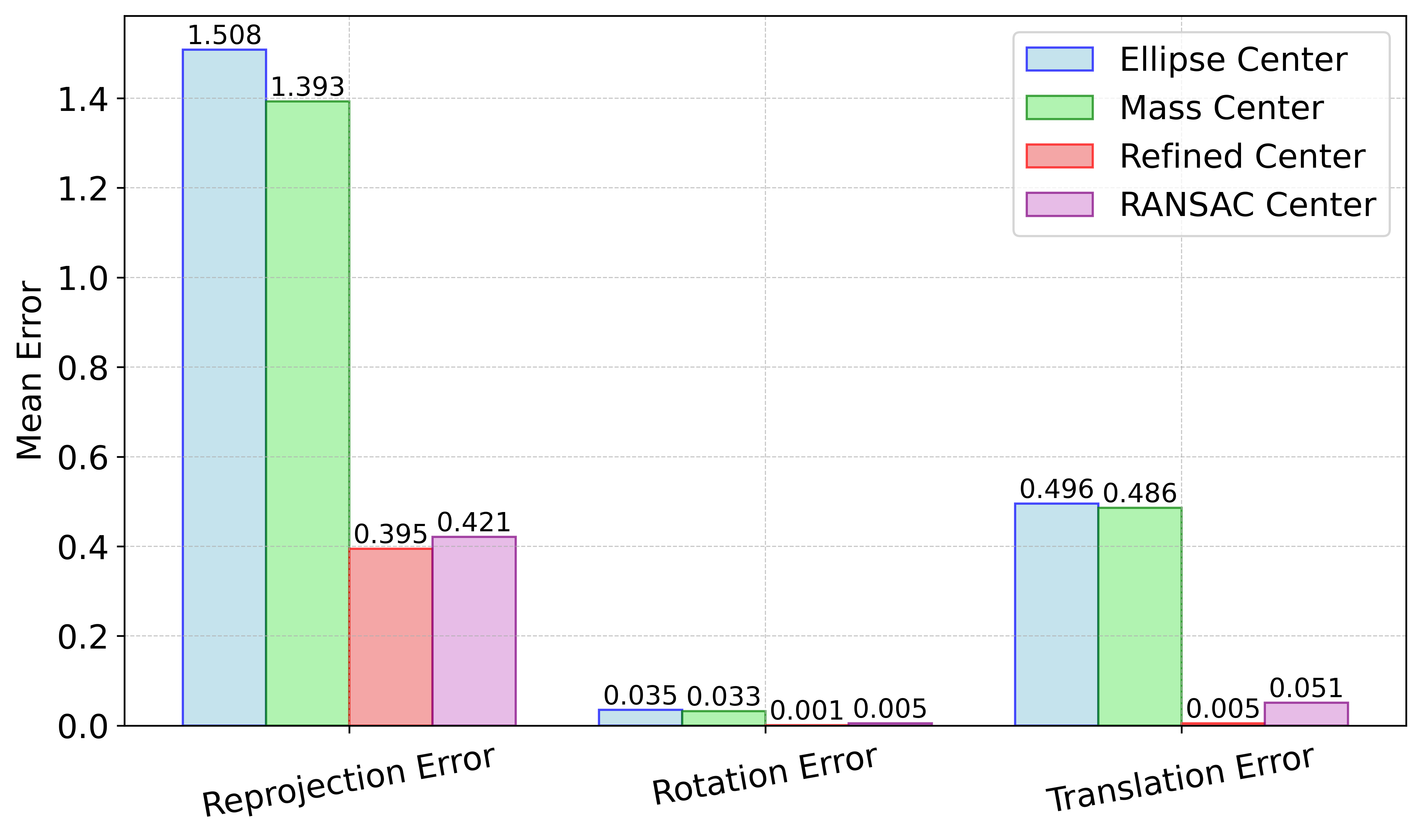}
    \caption{Effect of 2D center measurement on pose estimation.}
    \label{fig:2d_pose}
\end{figure}
\begin{figure*}[!htbp]
    \centering
    \includegraphics[width=0.98\textwidth]{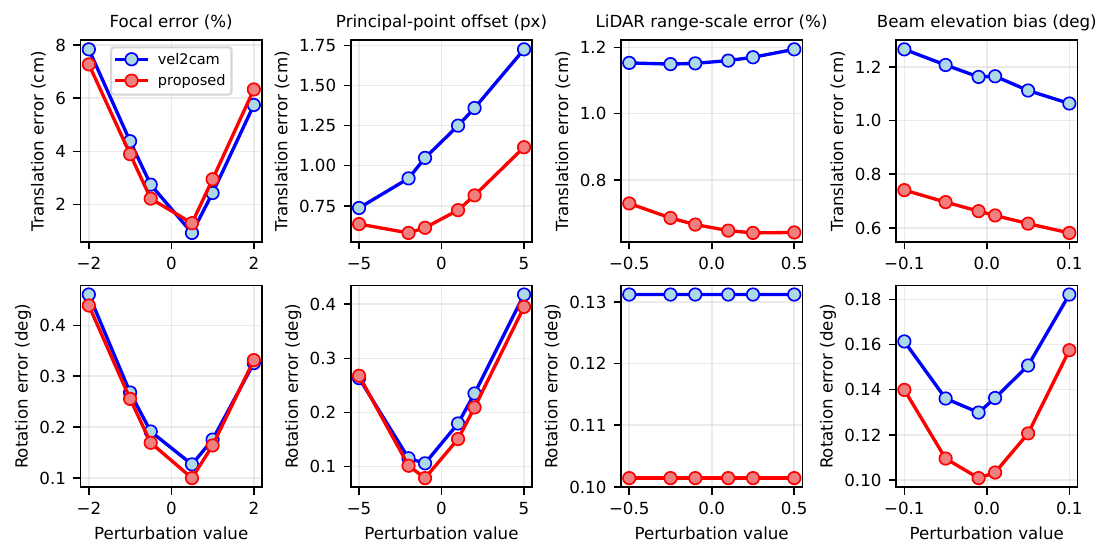}
    \caption{\revision{Gazebo K1 multi-pose sensitivity to camera focal/principal-point error and LiDAR range/beam-elevation error. Each point averages 30 paired seeds.}}
    \label{fig:intrinsic_sensitivity}
\end{figure*}

\begin{figure*}[!htbp]
    \centering
    \begin{subfigure}[b]{0.245\textwidth}
        \centering
        \includegraphics[width=\textwidth]{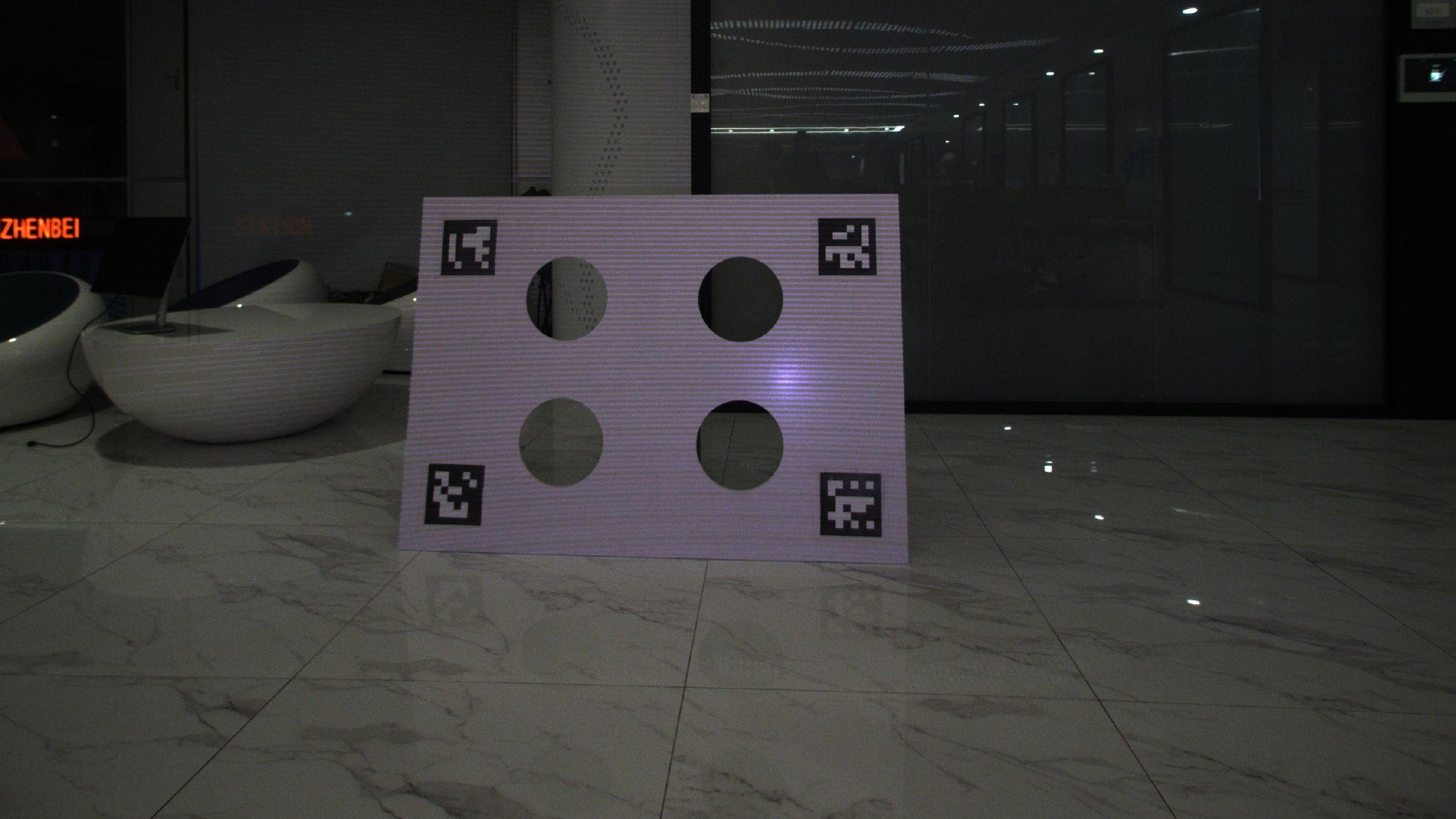}\\
        \includegraphics[width=\textwidth]{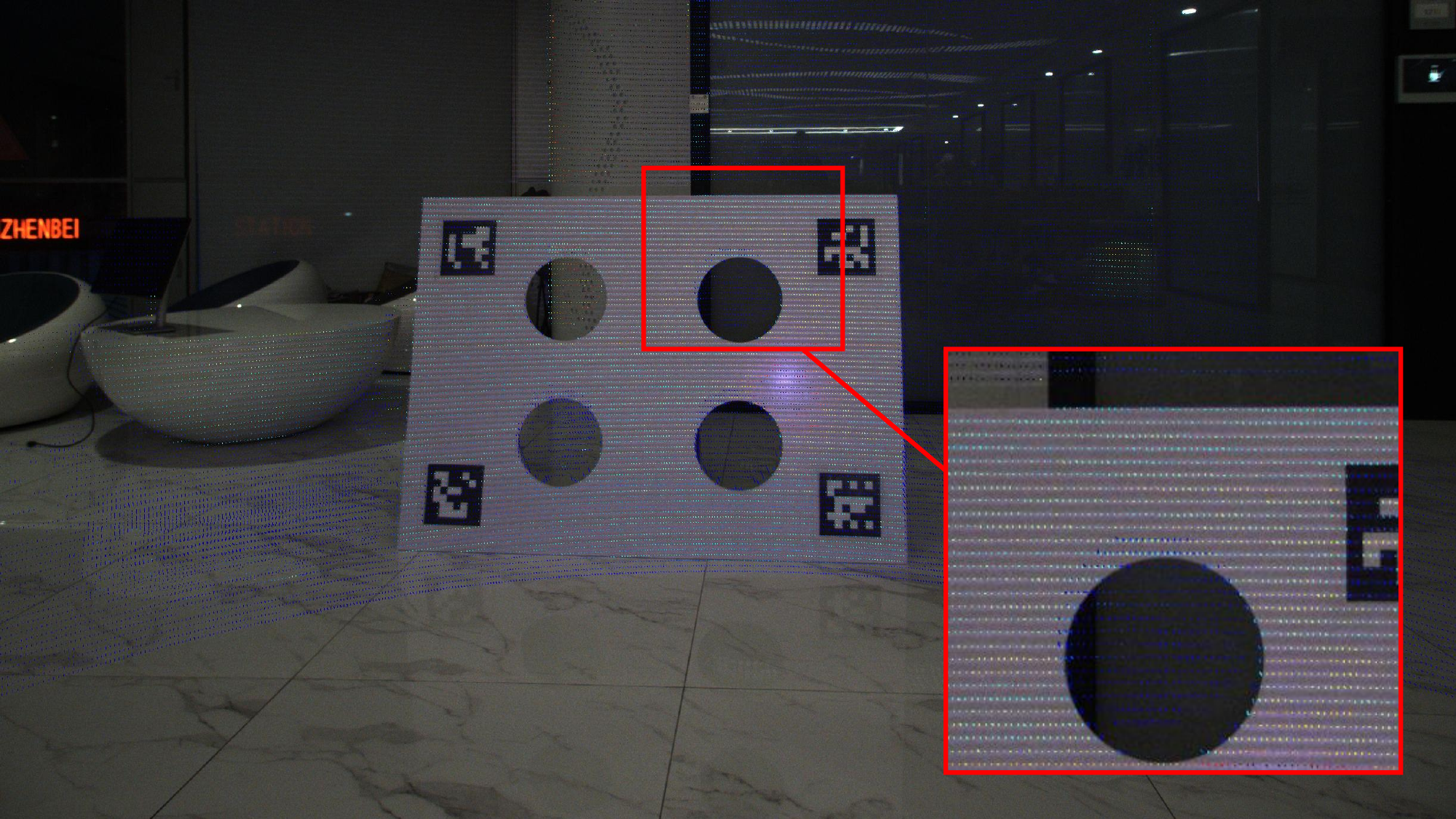}
        \caption{Planar board~\cite{zheng2025fast}}
        \label{fig:sub1}
    \end{subfigure}
    \hfill
    \begin{subfigure}[b]{0.245\textwidth}
        \centering
        \includegraphics[width=\textwidth]{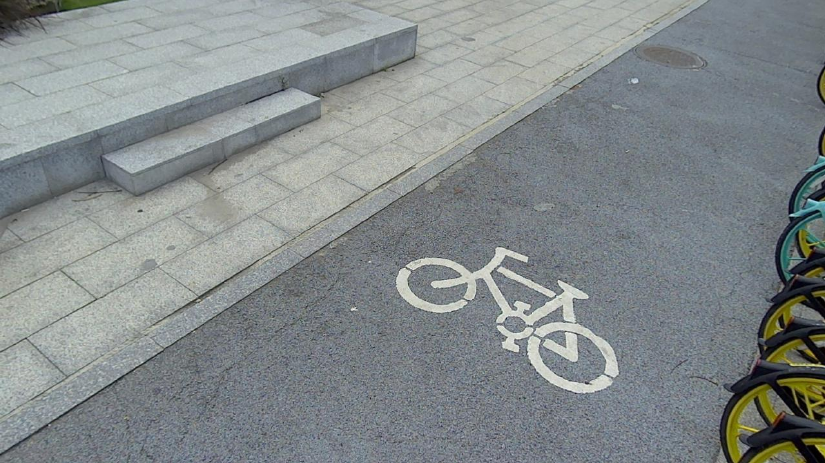}\\
        \includegraphics[width=\textwidth]{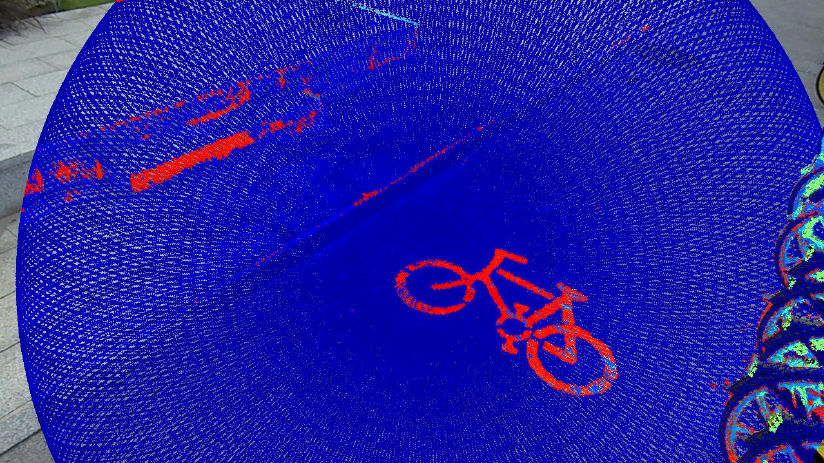}
        \caption{Natural target~\cite{chen2022pbacalib}}
        \label{fig:sub4}
    \end{subfigure}
    \hfill
    \begin{subfigure}[b]{0.245\textwidth}
        \centering
        \includegraphics[width=\textwidth]{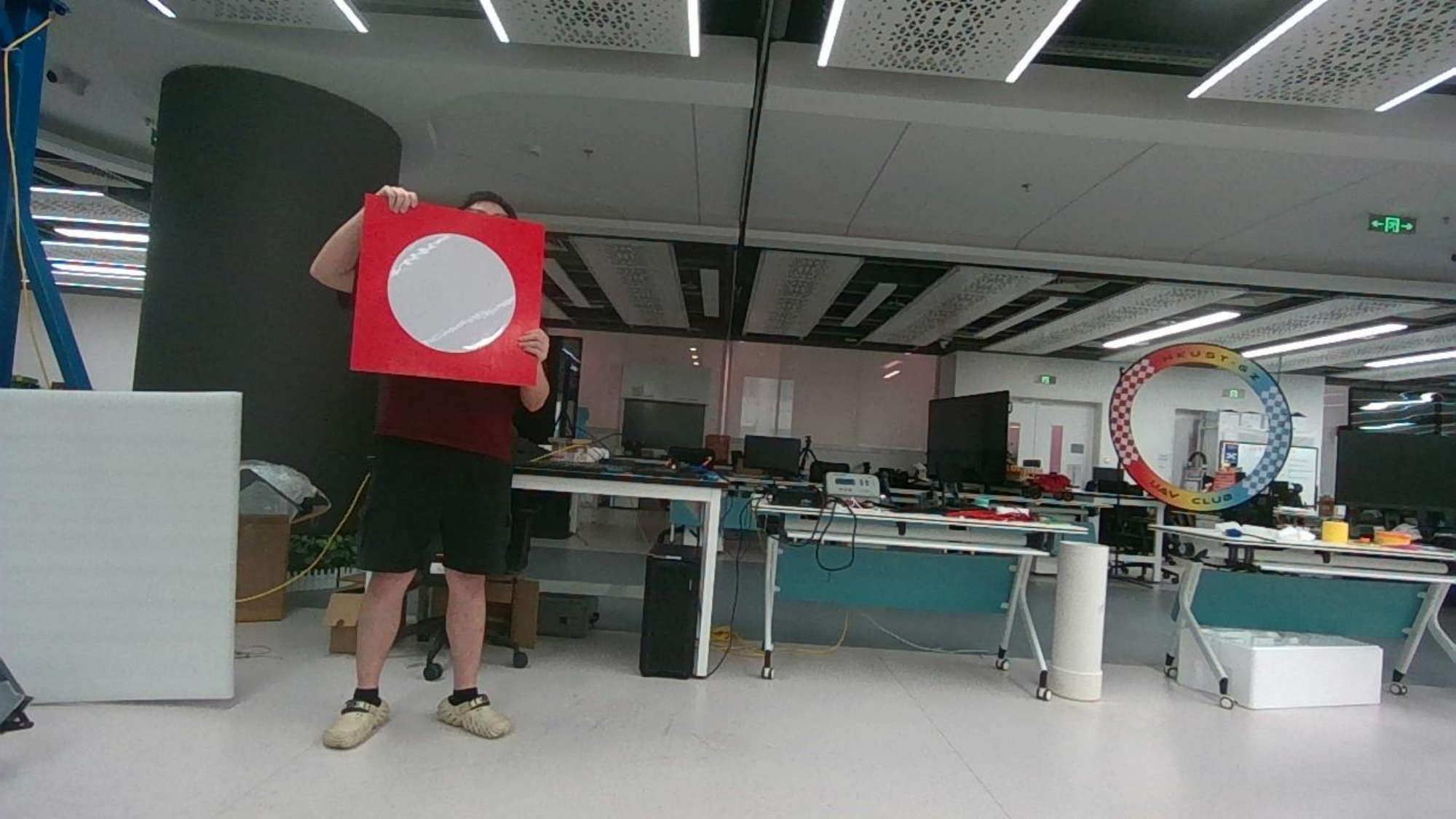}\\
        \includegraphics[width=\textwidth]{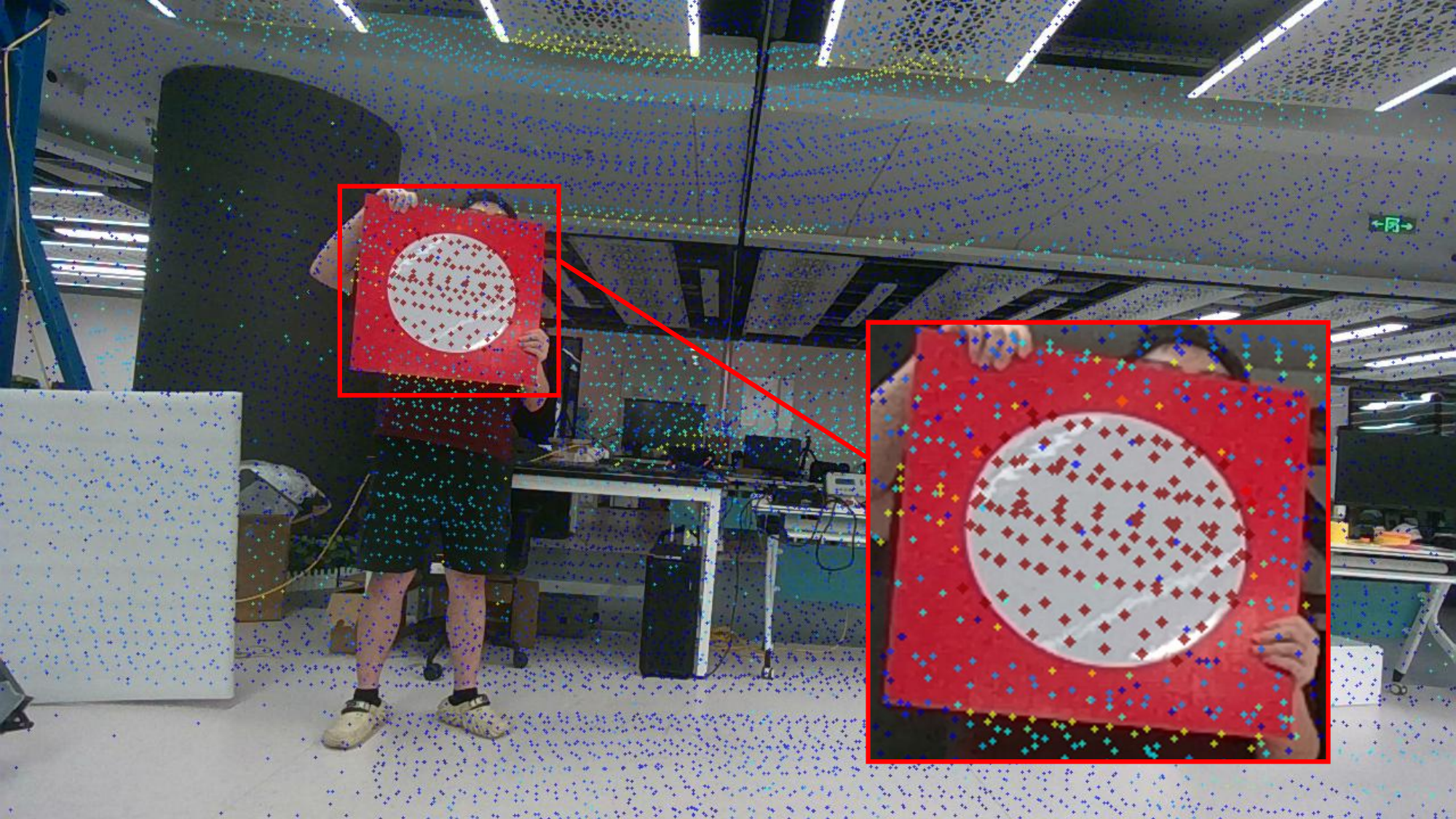}
        \caption{Setup 1}
        \label{fig:sub2}
    \end{subfigure}
    \hfill
    \begin{subfigure}[b]{0.245\textwidth}
        \centering
        \includegraphics[width=\textwidth]{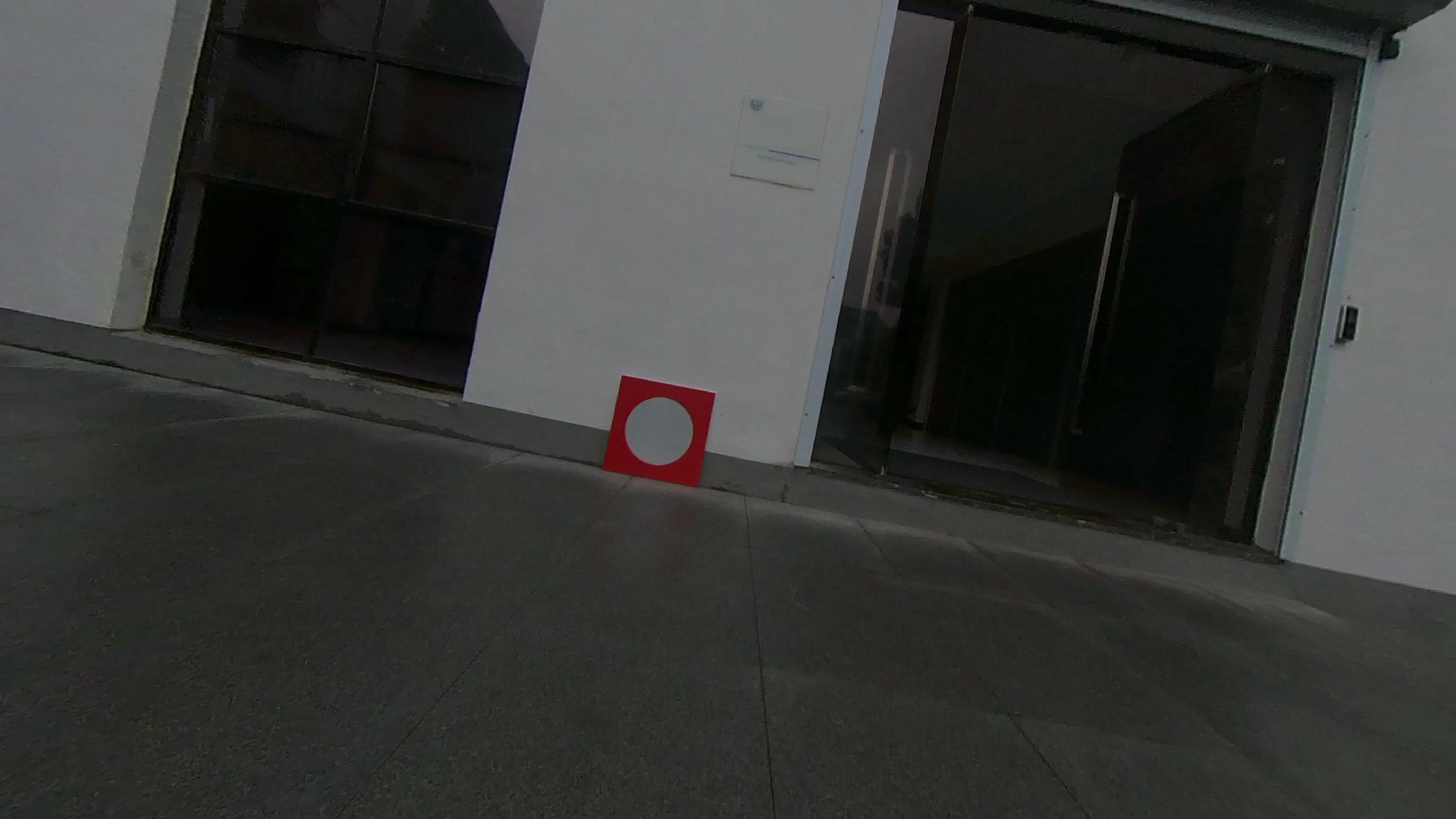}\\
        \includegraphics[width=\textwidth]{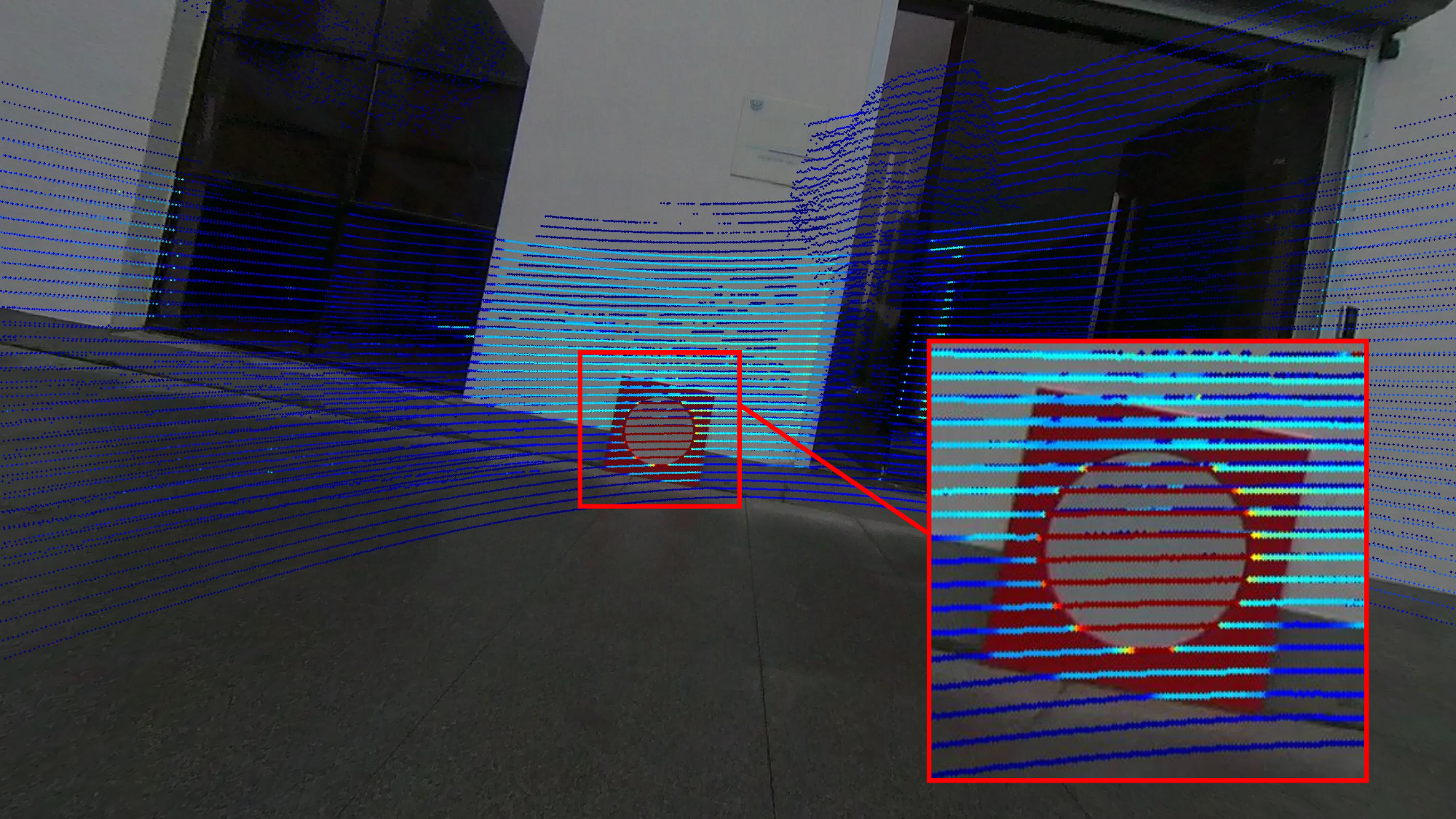}
        \caption{Setup 2}
        \label{fig:sub3}
    \end{subfigure}
\caption{Real-world LiDAR--camera calibration results. For each case, the LiDAR point cloud is projected into the image using the estimated extrinsics.}
    \label{fig:exp_real}
\end{figure*}

We further evaluate the influence of 2D center measurement on pose estimation. In each trial, 20 random 3D circle pairs are generated and PnP-RANSAC is used to estimate the camera pose from different 2D center measurements. Errors are reported using reprojection error, rotation error $\epsilon_{\mathrm{rot}}=\|\log({\mathbf{R}_{\mathrm{gt}}}^{-1}\mathbf{R}_{\mathrm{est}})^\vee\|$, and translation error $\epsilon_{\mathrm{trans}}=\|\mathbf{t}_{\mathrm{est}}-\mathbf{t}_{\mathrm{gt}}\|_2$. \Cref{fig:2d_pose} shows that the proposed 2D centers substantially reduce all three pose errors compared with the ellipse-center and center-of-mass baselines.

\subsubsection{\revision{Quasi-RANSAC Evaluation}}
\rev{
To evaluate the proposed quasi-RANSAC fallback, we design the following experiment. In each trial, $n\in\{8,12,20\}$ 3D points are sampled within a $2.4\times1.6\times3.0$~m volume at depths between 2 and 5~m. The points are projected using the camera model from the preceding 2D experiment. Gaussian noise with a standard deviation of 0.35 pixels is added to the correct image candidate. The false candidate is displaced in a random direction by a distance between 25 and 100 pixels. For an outlier correspondence, the two candidates are independently displaced by up to 150 pixels along each image coordinate. We evaluate outlier ratios of $\{0,0.1,0.2,0.3\}$ and confidence values of $\{0.95,0.99,0.999\}$. Each configuration is repeated 1000 times. A pose is considered successful when its rotation error is below $1^\circ$ and its translation error is below 5~cm. As can be seen from~\cref{tab:quasi}, without outliers, the observed success rate increases from 0.979 with eight correspondences to 0.995 with 20 correspondences. When the outlier ratio increases from 0.1 to 0.3, the success rate with eight correspondences decreases gradually, while with more correspondences, the success rate improves as expected. These results also clarify the meaning of~\eqref{eq:quasi_ransac_iterations}. The equation controls the probability of sampling at least one minimal set containing valid correspondences and the correct candidate branches. It does not directly determine the final pose success rate, which is also affected by image noise, the geometric conditioning of PnP, and consensus selection. Adaptive stopping increases the mean runtime for 20 correspondences from approximately 15~ms without outliers to 60~ms at an outlier ratio of 0.3. The method is therefore reliable when redundant correspondences are available. 
\begin{table}[!t]
    \caption{Quasi-RANSAC Planning Bound and Empirical Success}
    \label{tab:quasi}
    \centering
    \begin{tabular}{lrrrrr}
\toprule
Corr. & Outliers (\%) & Theory & Observed & Iterations & Time (ms) \\
\midrule
8 & 0 & 0.990 & 0.979 & 71.2 & 14.95 \\
8  & 10 & 0.990 & 0.948 & 123.8 & 25.22 \\
8 & 20 & 0.990 & 0.892 & 221.8 & 46.89 \\
8 & 30 & 0.990 & 0.878 & 220.8 & 46.78 \\
12 & 0 & 0.990 & 0.987 & 71.1 & 14.84 \\
12 & 10 & 0.990 & 0.981 & 103.0 & 21.06 \\
12 & 20 & 0.990 & 0.979 & 148.2 & 30.83 \\
12 & 30 & 0.990 & 0.918 & 358.3 & 72.54 \\
20 & 0 & 0.990 & 0.995 & 71.6 & 14.74 \\
20 & 10 & 0.990 & 0.989 & 110.0 & 22.24 \\
20 & 20 & 0.990 & 0.986 & 176.3 & 35.70 \\
20 & 30 & 0.990 & 0.974 & 301.3 & 60.06 \\
\bottomrule
\end{tabular}
\end{table}
}

\rev{
\subsection{Implementation and Computational Cost}
This section investigates the runtime of the proposed solution. The experiments were conducted on an Intel Xeon Gold 5218 server. The timing was performed for 1000 repetitions for all methods, and the results are presented in~\cref{tab:3druntime}. The proposed CGA has a p95 latency of 1.489~ms, and robust estimation raises it to 7.769~ms, which is under expectation as CGA is $O(n)$ and CGA-RANSAC is $O(In)$ for $I$ hypotheses as analyzed in~\cref{sec: 3dc}. PCL achieves roughly 0.1 ms per run thanks to its C++ implementation and possible parallelization. Nevertheless, all 3D variants remain well below 10~ms per target. The median peak Resident Set Size (RSS) values are 38.5, 40.0, and 37.6~MiB for CGA, CGA-RANSAC, and PCL, respectively.
}

\begin{table}[!htbp]
    \caption{\revision{3D Runtime and Process Memory}}
    \label{tab:3druntime}
    \centering
    \begin{tabular}{lrrr}
\toprule
\revision{Method ($n=64$) }& \revision{p50 (ms) }& \revision{p95 (ms) }& \revision{Peak RSS (MiB) }\\
\midrule
\revision{CGA }& \revision{1.407 }& \revision{1.489 }& \revision{38.5 }\\
\revision{CGA-RANSAC }& \revision{6.130 }& \revision{7.769 }& \revision{40.0 }\\
\revision{PCL SACMODEL }& \revision{0.098 }& \revision{0.100 }& \revision{37.6 }\\
\bottomrule
\end{tabular}
\end{table}

\subsection{\revision{Extrinsic Accuracy with Simulator Ground Truth}}
\rev{
Ground truth calibration parameters are typically unavailable, which is why Gazebo simulations are commonly used in the robotics and autonomous driving communities to evaluate accuracy in a controlled, fully observable environment. We therefore employ the Gazebo-based benchmark proposed by~\cite{Beltran} in ROS Noetic/Gazebo 11 with three noise levels: K0 (camera/LiDAR noise $0/0$), K1 ($0.007/0.008$~m), and K2 ($0.014/0.016$~m). 
Single-pose P1-P3 and multi-pose S1-S5 each contain 30 simulator seeds. The ground truth transform is read directly from the simulator sensor poses.
}

\begin{table}[!t]
    \caption{\revision{Gazebo Multi-Pose Accuracy Against Ground Truth (30 Seeds). Solution (Sol.) representation: velo2cam (v2c) and proposed (pp)}}
    \label{tab:gazebo_multi}
    \centering
    \begin{tabular}{llrrrr}
\toprule
 & \revision{Sol. }& \revision{3D (mm) }& \revision{Trans. (cm) }& \revision{Rot. (deg) }& \revision{Reproj. (px) }\\
\midrule
\multirow{2}{*}{K0} & \revision{v2c }& \revision{5.41$\pm$0.00 }& \revision{1.33$\pm$0.12 }& \revision{0.17$\pm$0.02 }& \revision{1.22$\pm$0.01 }\\
 &  \revision{pp }& \revision{3.37$\pm$0.00 }& \revision{0.55$\pm$0.06 }& \revision{0.10$\pm$0.00 }& \revision{0.73$\pm$0.04 }\\
\multirow{2}{*}{K1} & \revision{v2c  }& \revision{9.45$\pm$2.52 }& \revision{1.16$\pm$0.29 }& \revision{0.13$\pm$0.04 }& \revision{2.06$\pm$0.59 }\\
 & \revision{pp }& \revision{3.48$\pm$0.10 }& \revision{0.65$\pm$0.08 }& \revision{0.10$\pm$0.01 }& \revision{0.79$\pm$0.05 }\\
\multirow{2}{*}{K2} & \revision{v2c  }& \revision{22.06$\pm$5.24 }& \revision{1.13$\pm$0.50 }& \revision{0.18$\pm$0.06 }& \revision{4.85$\pm$1.33 }\\
 &  \revision{pp }& \revision{3.75$\pm$0.14 }& \revision{0.76$\pm$0.10 }& \revision{0.11$\pm$0.01 }& \revision{0.91$\pm$0.30 }\\
\bottomrule
\end{tabular}
\end{table}

\rev{\Cref{tab:gazebo_multi} compares the original \textit{velo2cam} pipeline (v2c) with the complete proposed pipeline (pp), and reports mean$\pm$standard deviation over all seeds. At the K1 noise level, the proposed method reduces translation error from 1.16 to 0.65~cm, rotation error from $0.13^\circ$ to $0.10^\circ$, and reprojection error from 2.06 to 0.79~px. The reduction persists across K0-K2, linking the center-level gains to improved end-to-end calibration.}

\rev{
To examine calibration-model sensitivity, camera focal length, principal point, LiDAR range scale, and LiDAR beam elevation are perturbed one at a time while the simulated extrinsic transform remains fixed. The tested magnitudes are $\pm\{0.5,1,2\}$\% for focal length, $\pm\{1,2,5\}$~px for the principal point, $\pm\{0.1,0.25,0.5\}$\% for range scale, and $\pm\{0.01,0.05,0.1\}^{\circ}$ for beam elevation. The 2D center is recomputed with each perturbed intrinsic matrix, whereas the range and elevation perturbations are applied in the LiDAR measurement frame before PnP. Every point in~\cref{fig:intrinsic_sensitivity} averages the same 30 paired K1 simulator seeds.} 
\rev{Focal-length error has the largest effect. At $\pm2$\%, both pipelines exhibit translation errors of approximately 6-8~cm and rotation errors of $0.32^\circ$-$0.46^\circ$. This result indicates that improved center estimation cannot compensate for an inaccurate projective camera model. Principal-point error causes a milder and asymmetric translation drift, which reaches 1.73~cm for \textit{velo2cam} and 1.11~cm for the proposed pipeline at $+5$~px. Within the tested range, LiDAR range-scale error changes the translation error by less than 0.1~cm and has almost no effect on the rotation error. Beam-elevation bias has a more noticeable effect on rotation, with the error increasing to approximately $0.18^\circ$ at $+0.1^\circ$. The proposed pipeline produces lower errors than the baseline in all considered scenarios. However, it is also observed that reliable extrinsic calibration still requires accurately calibrated sensor intrinsics, which is also expected.}

\subsection{Real-World Calibration}
\begin{figure}[!htbp]
    \centering
    \begin{subfigure}[b]{0.48\linewidth}
        \centering
        \includegraphics[width=\linewidth]{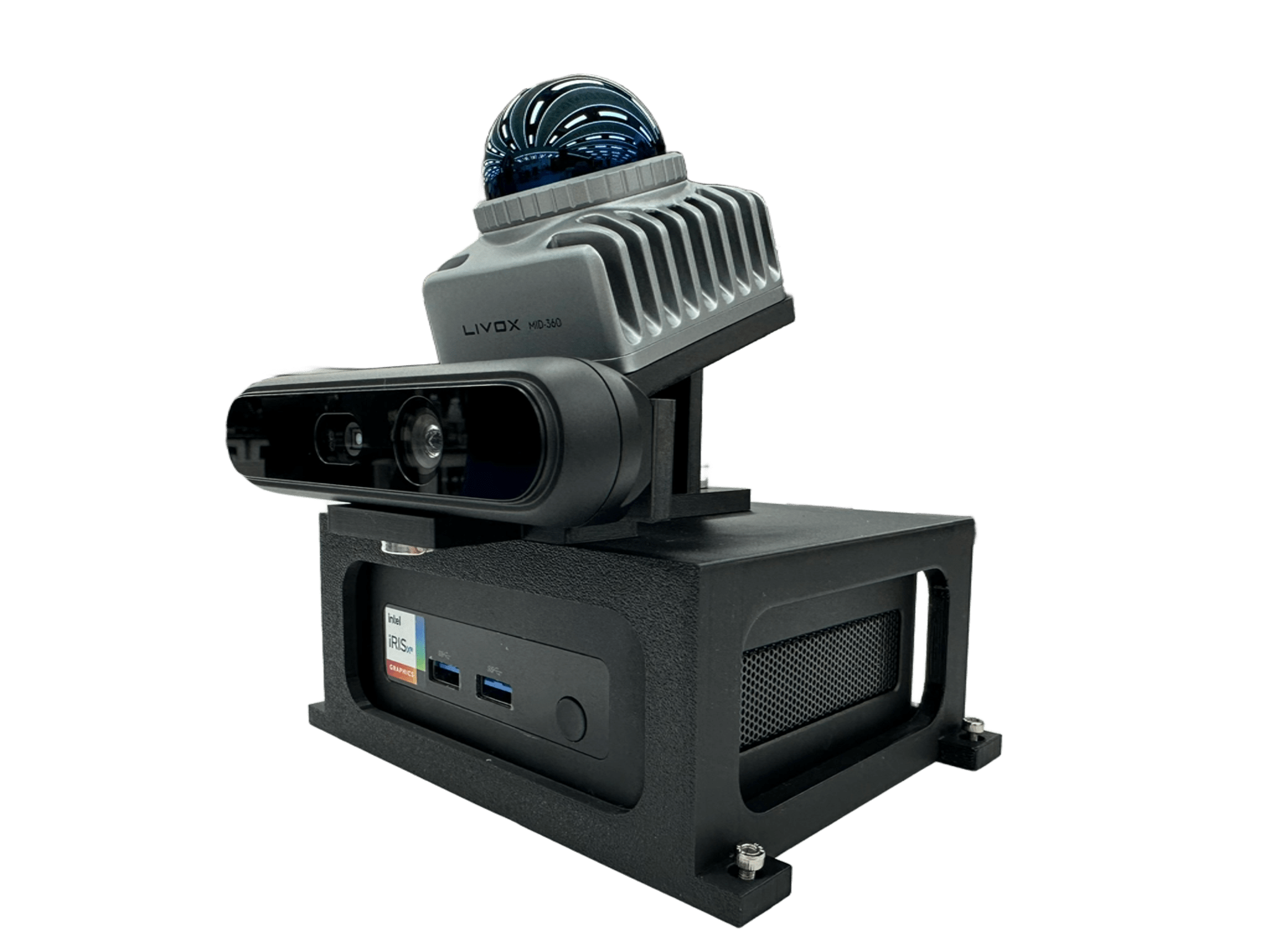}
        \caption{Orbbec camera and Livox Mid360 LiDAR.}
        \label{fig:setup_orbbec_livox}
    \end{subfigure}
    \hfill
    \begin{subfigure}[b]{0.48\linewidth}
        \centering
        \includegraphics[width=0.65\linewidth]{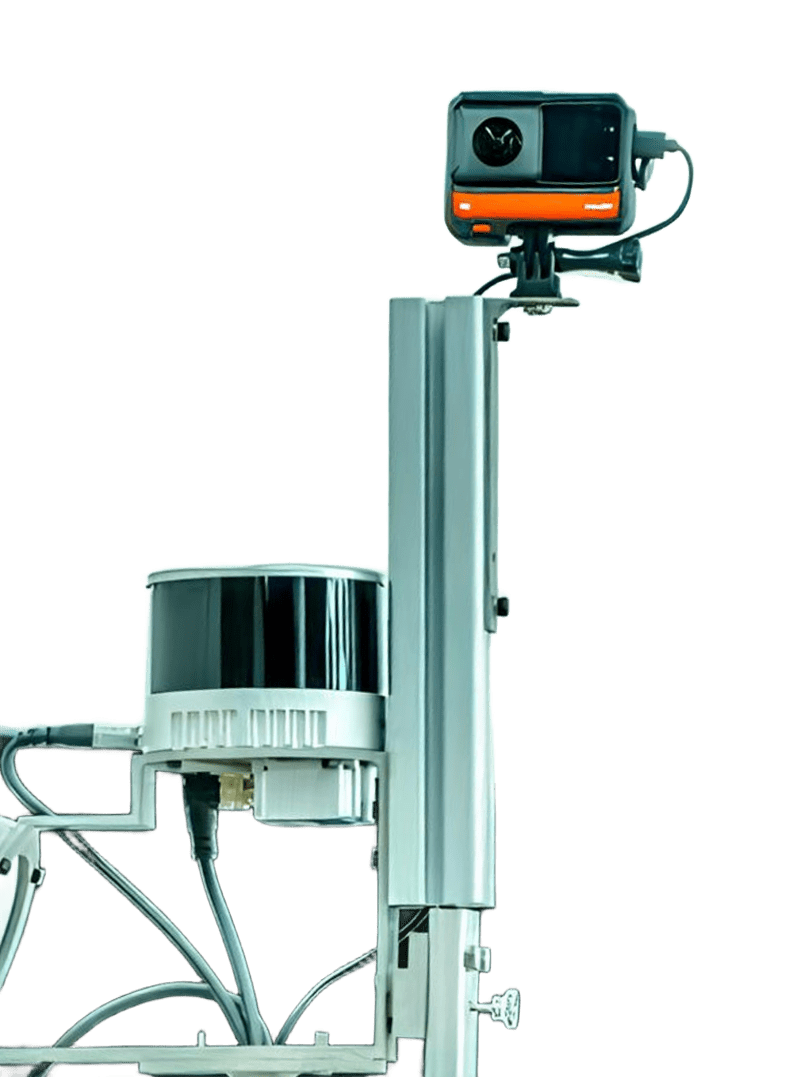}
        \caption{Insta360 camera and Hesai XT32 LiDAR.}
        \label{fig:setup_insta_hesai}
    \end{subfigure}
    \caption{Two custom sensor setups used for real-world validation.}
    \label{fig:custom_setups}
\end{figure}

Finally, we test real-world generality using two public datasets~\cite{zheng2025fast,chen2022pbacalib} and two custom-collected setups. The public data include planar circular targets with multiple LiDAR--camera configurations and a natural bicycle-wheel target. The custom setups cover a pinhole camera with a solid-state LiDAR and a fisheye camera with a mechanical LiDAR, as shown in~\cref{fig:custom_setups}. Since ground-truth extrinsics are unavailable, we validate the results by projecting LiDAR points into the image. The examples in~\cref{fig:exp_real} show consistent point-image alignment across target types and sensor modalities, supporting the applicability of the proposed circular-center measurement strategy to heterogeneous \revision{LiDAR-camera }systems.

\section{Conclusions}
\label{sec: conclusion}
This paper studied the accurate measurement of 3D and 2D circular centers, with LiDAR-camera extrinsic calibration used as a representative application. We showed that conventional circular-target pipelines can suffer from two systematic measurement biases: decoupled 3D plane-circle fitting in LiDAR and the incorrect use of the image ellipse center as the projection of the physical 3D circle center. To reduce these errors, we proposed a CGA-RANSAC estimator for robust 3D circular-center measurement and a chord-length-variance method for perspective-aware 2D projected-center measurement. The ambiguity of the 2D loss was handled by homography validation when coplanar circles are available and by a quasi-RANSAC fallback otherwise.

Experiments on numerical simulations, synthetic calibration data, and real-world sensor setups verified that more accurate 3D and 2D center measurements lead to lower extrinsic calibration errors and more consistent point-image alignment. The results support the use of circular-center measurement as a practical geometric primitive for heterogeneous sensing systems. Future work will focus on sub-pixel 2D optimization, formal propagation of center-measurement errors to downstream parameters, and broader validation on additional sensor configurations and circular-feature measurement tasks.

\section*{Acknowledgments}
The authors thank Prof. Yu Zhang, Changjian Jiang, Zeyu Wan, and Yiming Zhu for their support in data collection and experiments.

\bibliographystyle{IEEEtran}
\bibliography{done/circle,done/dl,done/ellipse,done/lib,done/method,done/ref,done/tim_im_refs}

@ARTICLE{zheng2025fast,
  author={Zheng, Chunran and Zhang, Fu},
  journal={IEEE Robotics and Automation Practice}, 
  title={FAST-Calib: LiDAR-Camera Extrinsic Calibration in One Second}, 
  year={2026},
  volume={1},
  number={},
  pages={108-112},
  doi={10.1109/RAP.2026.3692446}}

@article{Fremont,
author = {Fremont, Vincent and Rodriguez Florez, Sergio and Bonnifait, Philippe},
year = {2012},
month = {12},
pages = {1-27},
title = {Circular Targets for 3D Alignment of Video and Lidar Sensors},
volume = {0},
journal = {Advanced Robotics},
doi = {10.1080/01691864.2012.703235}
}

@INPROCEEDINGS{opencalib,
  author={Yan, Guohang and He, Feiyu and Shi, Chunlei and Wei, Pengjin and Cai, Xinyu and Li, Yikang},
  booktitle={2023 IEEE International Conference on Robotics and Automation (ICRA)}, 
  title={Joint Camera Intrinsic and LiDAR-Camera Extrinsic Calibration}, 
  year={2023},
  volume={},
  number={},
  pages={11446-11452},
  doi={10.1109/ICRA48891.2023.10160542}}

@ARTICLE{KimRAL,
  author={Kim, Daeho and Shin, Seunghui and Hwang, Hyoseok},
  journal={IEEE Robotics and Automation Letters}, 
  title={Camera-LiDAR Extrinsic Calibration Using Constrained Optimization With Circle Placement}, 
  year={2025},
  volume={10},
  number={2},
  pages={883-890},
  doi={10.1109/LRA.2024.3512253}}

@ARTICLE{Beltran,
  author={Beltrán, Jorge and Guindel, Carlos and de la Escalera, Arturo and García, Fernando},
  journal={IEEE Transactions on Intelligent Transportation Systems}, 
  title={Automatic Extrinsic Calibration Method for LiDAR and Camera Sensor Setups}, 
  year={2022},
  volume={23},
  number={10},
  pages={17677-17689},
  doi={10.1109/TITS.2022.3155228}}

@INPROCEEDINGS{Kummerle,
  author={Kümmerle, Julius and Kühner, Tilman and Lauer, Martin},
  booktitle={2018 IEEE/RSJ International Conference on Intelligent Robots and Systems (IROS)}, 
  title={Automatic Calibration of Multiple Cameras and Depth Sensors with a Spherical Target}, 
  year={2018},
  volume={},
  number={},
  pages={1-8},
  doi={10.1109/IROS.2018.8593955}}

@ARTICLE{ZhangRAL,
  author={Zhang, Guanyu and Wu, Kunyang and Lin, Jun and Wang, Tianhao and Liu, Yang},
  journal={IEEE Robotics and Automation Letters}, 
  title={Automatic Extrinsic Parameter Calibration for Camera-LiDAR Fusion Using Spherical Target}, 
  year={2024},
  volume={9},
  number={6},
  pages={5743-5750},
  doi={10.1109/LRA.2024.3397072}}

@ARTICLE{tim_full_2025,
  author={Wei, Ming and Lu, Jun-Guo and Ye, Nan and Zhu, Zhen and Zhang, Qinghao and Wang, Yafei},
  journal={IEEE Transactions on Instrumentation and Measurement}, 
  title={A Full-Lifecycle Calibration Method for Camera and LiDAR in Autonomous Driving}, 
  year={2025},
  volume={74},
  number={},
  pages={1-17},
  doi={10.1109/TIM.2025.3565783}
}

@ARTICLE{liu_sensors_2023,
  author={Liu, Haitao and Xu, Qingpo and Huang, Yugeng and Ding, Yabin and Xiao, Juliang},
  journal={IEEE Sensors Journal}, 
  title={A Method for Synchronous Automated Extrinsic Calibration of LiDAR and Cameras Based on a Circular Calibration Board}, 
  year={2023},
  volume={23},
  number={20},
  pages={25026-25035},
  doi={10.1109/JSEN.2023.3312322}}

@ARTICLE{A4LidarTag,
  author={Xie, Yusen and Deng, Lei and Sun, Ting and Fu, Yeyu and Li, Jian and Cui, Xinglong and Yin, Hanxi and Deng, Shuixin and Xiao, Junwei and Chen, Baohua},
  journal={IEEE Robotics and Automation Letters}, 
  title={A4LidarTag: Depth-Based Fiducial Marker for Extrinsic Calibration of Solid-State Lidar and Camera}, 
  year={2022},
  volume={7},
  number={3},
  pages={6487-6494},
  doi={10.1109/LRA.2022.3173033}}

@INPROCEEDINGS{Domhoficra,
  author={Domhof, Joris and Kooij, Julian F.P. and Gavrila, Dariu M.},
  booktitle={2019 International Conference on Robotics and Automation (ICRA)}, 
  title={An Extrinsic Calibration Tool for Radar, Camera and Lidar}, 
  year={2019},
  volume={},
  number={},
  pages={8107-8113},
  doi={10.1109/ICRA.2019.8794186}}

@ARTICLE{Xingxing,
  author={Li, Xingxing and He, Feiyang and Li, Shengyu and Zhou, Yuxuan and Xia, Chunxi and Wang, Xuanbin},
  journal={IEEE Sensors Journal}, 
  title={Accurate and Automatic Extrinsic Calibration for a Monocular Camera and Heterogenous 3D LiDARs}, 
  year={2022},
  volume={22},
  number={16},
  pages={16472-16480},
  doi={10.1109/JSEN.2022.3189041}}

@ARTICLE{CMRNext,
  author={Cattaneo, Daniele and Valada, Abhinav},
  journal={IEEE Transactions on Robotics}, 
  title={CMRNext: Camera to LiDAR Matching in the Wild for Localization and Extrinsic Calibration}, 
  year={2025},
  volume={41},
  number={},
  pages={1995-2013},
  doi={10.1109/TRO.2025.3546784}}

@INPROCEEDINGS{SGCalib,
  author={Lin, Zhipeng and Gao, Zhi and Liu, Xinyi and Wang, Jialiang and Song, Weiwei and Chen, Ben M. and Li, Chenyang and Huang, Yue and Zhu, Yuhan},
  booktitle={2024 IEEE International Conference on Robotics and Automation (ICRA)}, 
  title={SGCalib: A Two-stage Camera-LiDAR Calibration Method Using Semantic Information and Geometric Features}, 
  year={2024},
  volume={},
  number={},
  pages={14527-14533},
  doi={10.1109/ICRA57147.2024.10610560}}

@INPROCEEDINGS{3DGScalib,
  author={Herau, Quentin and Bennehar, Moussab and Moreau, Arthur and Piasco, Nathan and Roldão, Luis and Tsishkou, Dzmitry and Migniot, Cyrille and Vasseur, Pascal and Demonceaux, Cédric},
  booktitle={2024 IEEE/RSJ International Conference on Intelligent Robots and Systems (IROS)}, 
  title={3DGS-Calib: 3D Gaussian Splatting for Multimodal SpatioTemporal Calibration}, 
  year={2024},
  volume={},
  number={},
  pages={8315-8321},
  doi={10.1109/IROS58592.2024.10801360}}

@inproceedings{elsd,
  title={A parameterless line segment and elliptical arc detector with enhanced ellipse fitting},
  author={P{\u{a}}tr{\u{a}}ucean, Viorica and Gurdjos, Pierre and Von Gioi, Rafael Grompone},
  booktitle={Eur. Conf. Comput. Vis.},
  pages={572--585},
  year={2012},
  organization={Springer}
}

@article{aamed,
  title={Arc adjacency matrix-based fast ellipse detection},
  author={Meng, Cai and Li, Zhaoxi and Bai, Xiangzhi and Zhou, Fugen},
  journal={IEEE Trans. Image Process.},
  volume={29},
  pages={4406--4420},
  year={2020},
  publisher={IEEE}
}

@article{lu2019arc,
  title={Arc-support line segments revisited: An efficient high-quality ellipse detection},
  author={Lu, Changsheng and Xia, Siyu and Shao, Ming and Fu, Yun},
  journal={IEEE Trans. Image Process.},
  volume={29},
  pages={768--781},
  year={2019},
}

@article{RHT,
 title={Randomized Hough transform: improved ellipse detection with comparison},
 author={McLaughlin, Robert A},
 journal={Pattern Recognit. Lett.},
 volume={19},
 number={3-4},
 pages={299--305},
 year={1998},
 }

@InProceedings{Rusu_ICRA2011_PCL,
  author    = {Radu Bogdan Rusu and Steve Cousins},
  title     = {{3D is here: Point Cloud Library (PCL)}},
  booktitle = {{IEEE International Conference on Robotics and Automation (ICRA)}},
  month     = {May 9-13},
  year      = {2011},
  address   = {Shanghai, China},
  publisher = {IEEE}
}

@article{opencv_library,
    author = {Bradski, G.},
    journal = {Dr. Dobb's Journal of Software Tools},
    title = {{The OpenCV Library}},
    year = {2000}
}

@article{cucci2016accurate,
  title={Accurate optical target pose determination for applications in aerial photogrammetry},
  author={Cucci, DA},
  journal={ISPRS Annals of the Photogrammetry, Remote Sensing and Spatial Information Sciences},
  volume={3},
  pages={257--262},
  year={2016},
  publisher={Copernicus GmbH}
}

@ARTICLE{KimPAMI,
  author={Jun-Sik Kim and Gurdjos, P. and In-So Kweon},
  journal={IEEE Transactions on Pattern Analysis and Machine Intelligence}, 
  title={Geometric and algebraic constraints of projected concentric circles and their applications to camera calibration}, 
  year={2005},
  volume={27},
  number={4},
  pages={637-642},
  doi={10.1109/TPAMI.2005.80}}

@INPROCEEDINGS{cctag,
  author={Calvet, Lilian and Gurdjos, Pierre and Griwodz, Carsten and Gasparini, Simone},
  booktitle={2016 IEEE Conference on Computer Vision and Pattern Recognition (CVPR)}, 
  title={Detection and Accurate Localization of Circular Fiducials under Highly Challenging Conditions}, 
  year={2016},
  volume={},
  number={},
  pages={562-570},
  doi={10.1109/CVPR.2016.67}}

@INPROCEEDINGS{HONGBO,
author = {Hongbo, Li and David, Hestenes and Alyn Rockwood},
year = {2001},
month = {06},
pages = {27-59},
title = {Generalized Homogeneous Coordinates for Computational Geometry},
booktitle={Geometric Computing with Clifford Algebras},
publisher={Springer}
}

@ARTICLE{Hartley,
  author={Hartley, R.I.},
  journal={IEEE Transactions on Pattern Analysis and Machine Intelligence}, 
  title={In defense of the eight-point algorithm}, 
  year={1997},
  volume={19},
  number={6},
  pages={580-593},
  doi={10.1109/34.601246}}

@ARTICLE{Nister,
  author={Nister, D.},
  journal={IEEE Transactions on Pattern Analysis and Machine Intelligence}, 
  title={An efficient solution to the five-point relative pose problem}, 
  year={2004},
  volume={26},
  number={6},
  pages={756-770},
  doi={10.1109/TPAMI.2004.17}}

@article{Fischler1981RandomSC,
  title={Random sample consensus: a paradigm for model fitting with applications to image analysis and automated cartography},
  author={Martin A. Fischler and Robert C. Bolles},
  journal={Commun. ACM},
  year={1981},
  volume={24},
  pages={381-395},
  url={https://api.semanticscholar.org/CorpusID:972888}
}

@article{dorst,
author = {Dorst, Leo},
year = {2014},
month = {11},
pages = {},
title = {Total Least Squares Fitting of k-Spheres in n-D Euclidean Space Using an (n+2)-D Isometric Representation},
volume = {50},
journal = {Journal of Mathematical Imaging and Vision},
doi = {10.1007/s10851-014-0495-2}
}

@article{chen2022pbacalib,
  title={PBACalib: Targetless extrinsic calibration for high-resolution LiDAR-camera system based on plane-constrained bundle adjustment},
  author={Chen, Feiyi and Li, Liang and Zhang, Shuyang and Wu, Jin and Wang, Lujia},
  journal={IEEE Robotics and Automation Letters},
  volume={8},
  number={1},
  pages={304--311},
  year={2022},
  publisher={IEEE}
}

@inproceedings{geiger2012automatic,
  title={Automatic camera and range sensor calibration using a single shot},
  author={Geiger, Andreas and Moosmann, Frank and Car, {\"O}mer and Schuster, Bernhard},
  booktitle={2012 IEEE international conference on robotics and automation},
  pages={3936--3943},
  year={2012},
  organization={IEEE}
}

@article{yuan2021pixel,
  title={Pixel-level extrinsic self calibration of high resolution lidar and camera in targetless environments},
  author={Yuan, Chongjian and Liu, Xiyuan and Hong, Xiaoping and Zhang, Fu},
  journal={IEEE Robotics and Automation Letters},
  volume={6},
  number={4},
  pages={7517--7524},
  year={2021},
  publisher={IEEE}
}

@article{Pandey, title={Automatic Targetless Extrinsic Calibration of a 3D Lidar and Camera by Maximizing Mutual Information}, volume={26},  DOI={10.1609/aaai.v26i1.8379}, number={1}, journal={Proceedings of the AAAI Conference on Artificial Intelligence}, author={Pandey, Gaurav and McBride, James and Savarese, Silvio and Eustice, Ryan}, year={2021}, month={Sep.}, pages={2053-2059} }

@inproceedings{levinson2013automatic,
  title={Automatic online calibration of cameras and lasers.},
  author={Levinson, Jesse and Thrun, Sebastian},
  booktitle={Robotics: science and systems},
  volume={2},
  number={7},
  year={2013},
  organization={Berlin, Germany}
}

@ARTICLE{TaylorTRO,
  author={Taylor, Zachary and Nieto, Juan},
  journal={IEEE Transactions on Robotics}, 
  title={Motion-Based Calibration of Multimodal Sensor Extrinsics and Timing Offset Estimation}, 
  year={2016},
  volume={32},
  number={5},
  pages={1215-1229},
  doi={10.1109/TRO.2016.2596771}}

@ARTICLE{tim_ye_2022_rgkcnet,
  author={Ye, Chao and Pan, Huihui and Gao, Huijun},
  journal={IEEE Transactions on Instrumentation and Measurement}, 
  title={Keypoint-Based LiDAR-Camera Online Calibration With Robust Geometric Network}, 
  year={2022},
  volume={71},
  number={},
  pages={1-11},
  doi={10.1109/TIM.2021.3129882}}

@ARTICLE{tim_liu_2022_adaptive_voxel,
  author={Liu, Xiyuan and Yuan, Chongjian and Zhang, Fu},
  journal={IEEE Transactions on Instrumentation and Measurement}, 
  title={Targetless Extrinsic Calibration of Multiple Small FoV LiDARs and Cameras Using Adaptive Voxelization}, 
  year={2022},
  volume={71},
  number={},
  pages={1-12},
  doi={10.1109/TIM.2022.3176889}}

@ARTICLE{tim_zhang_2023_overlapfree,
  author={Zhang, Dedong and Ma, Lingfei and Gong, Zheng and Tan, Weikai and Zelek, John and Li, Jonathan},
  journal={IEEE Transactions on Instrumentation and Measurement}, 
  title={An Overlap-Free Calibration Method for LiDAR-Camera Platforms Based on Environmental Perception}, 
  year={2023},
  volume={72},
  number={},
  pages={1-7},
  doi={10.1109/TIM.2023.3244816}}

@ARTICLE{tim_cui_2024_aclc,
  author={Cui, Jiahe and Niu, Jianwei and He, Yunxiang and Liu, Dian and Ouyang, Zhenchao},
  journal={IEEE Transactions on Instrumentation and Measurement}, 
  title={ACLC: Automatic Calibration for Nonrepetitive Scanning LiDAR-Camera System Based on Point Cloud Noise Optimization}, 
  year={2024},
  volume={73},
  number={},
  pages={1-14},
  doi={10.1109/TIM.2023.3336753}}

@ARTICLE{tim_gong_2024_sceneaware,
  author={Gong, Zheng and He, Rui and Gao, Kyle and Cai, Guorong},
  journal={IEEE Transactions on Instrumentation and Measurement}, 
  title={Scene-Aware Online Calibration of LiDAR and Cameras for Driving Systems}, 
  year={2024},
  volume={73},
  number={},
  pages={1-9},
  doi={10.1109/TIM.2023.3342241}}

@ARTICLE{tim_zhao_2024_vanishing,
  author={Zhao, Yilin and Zhao, Long and Yang, Fengli and Li, Wangfang and Sun, Yi},
  journal={IEEE Transactions on Instrumentation and Measurement}, 
  title={Extrinsic Calibration of High Resolution LiDAR and Camera Based on Vanishing Points}, 
  year={2024},
  volume={73},
  number={},
  pages={1-12},
  doi={10.1109/TIM.2024.3450106}}

@ARTICLE{tim_liu_2024_transformer,
  author={Liu, Fengyu and Cao, Yi and Cheng, Xianghong and Wu, Xinyi},
  journal={IEEE Transactions on Instrumentation and Measurement}, 
  title={Transformer-Based Local-to-Global LiDAR-Camera Targetless Calibration With Multiple Constraints}, 
  year={2024},
  volume={73},
  number={},
  pages={1-13},
  doi={10.1109/TIM.2024.3420358}}

@ARTICLE{tim_wei_2025_lifecycle,
  author={Wei, Ming and Lu, Jun-Guo and Ye, Nan and Zhu, Zhen and Zhang, Qinghao and Wang, Yafei},
  journal={IEEE Transactions on Instrumentation and Measurement}, 
  title={A Full-Lifecycle Calibration Method for Camera and LiDAR in Autonomous Driving}, 
  year={2025},
  volume={74},
  number={},
  pages={1-17},
  doi={10.1109/TIM.2025.3565783}}

@article{wang2022pose,
  title={Pose error analysis method based on a single circular feature},
  author={Wang, Zepeng and Chen, Derong and Gong, Jiulu},
  journal={Pattern Recognition},
  volume={129},
  pages={108726},
  year={2022},
  publisher={Elsevier}
}

@article{hedstrand2024photogrammetry,
  title={Improving Photogrammetry Instrument Performance Through Camera Calibration for Precision Digital Manufacturing},
  author={Hedstrand, Therese and Southon, Nicholas and Martin, Oliver and Davey, Craig and Yu, Nan},
  journal={Procedia CIRP},
  volume={121},
  pages={91--96},
  year={2024},
  publisher={Elsevier}
}

@ARTICLE{scaramuzza2011vo,
  author={Scaramuzza, Davide and Fraundorfer, Friedrich},
  journal={IEEE Robotics \& Automation Magazine}, 
  title={Visual Odometry [Tutorial]}, 
  year={2011},
  volume={18},
  number={4},
  pages={80-92},
  doi={10.1109/MRA.2011.943233}}

@misc{wei2022circlenet,
      title={Deep Algebraic Fitting for Multiple Circle Primitives Extraction from Raw Point Clouds}, 
      author={Zeyong Wei and Honghua Chen and Hao Tang and Qian Xie and Mingqiang Wei and Jun Wang},
      year={2022},
      eprint={2204.00920},
      archivePrefix={arXiv},
      primaryClass={cs.CV},
      url={https://arxiv.org/abs/2204.00920}, 
}

@INPROCEEDINGS{song2024distorted,
  author={Song, Chaehyeon and Shin, Jaeho and Jeon, Myung-Hwan and Lim, Jongwoo and Kim, Ayoung},
  booktitle={2024 IEEE/CVF Conference on Computer Vision and Pattern Recognition (CVPR)}, 
  title={Unbiased Estimator for Distorted Conics in Camera Calibration}, 
  year={2024},
  volume={},
  number={},
  pages={373-381},
  doi={10.1109/CVPR52733.2024.00043}}

 



\vspace{11pt}
\vspace{-33pt}
\begin{IEEEbiography}[{\includegraphics[width=1in,height=1.25in,clip,keepaspectratio]{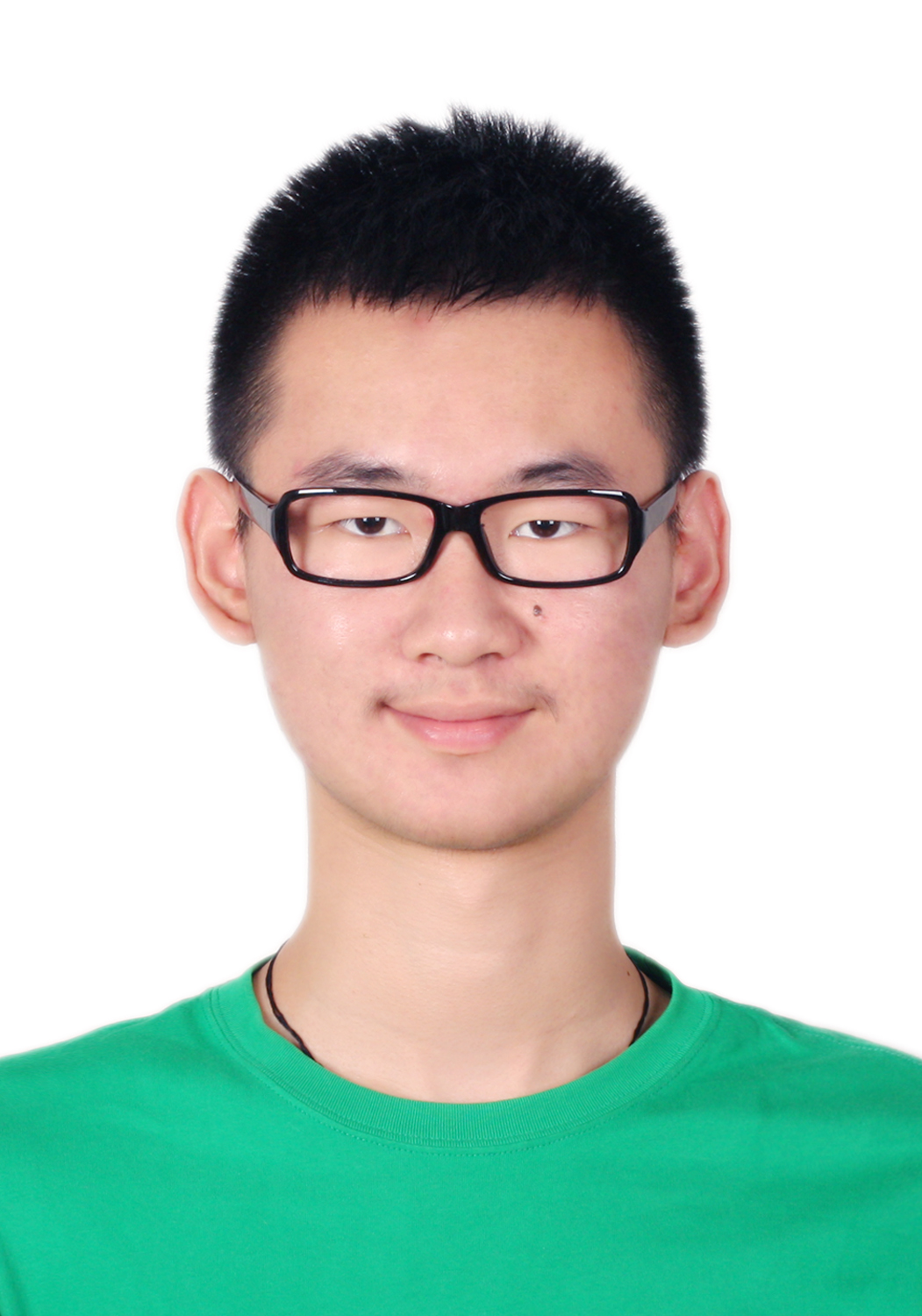}}]
{Jiajun Jiang} received the B.S. and M.S. degrees in control science and technology from Zhejiang University, Hangzhou, China, in 2019 and 2022, respectively. He formerly served as an Algorithm Engineer at Alibaba DAMO Academy. He is currently pursuing the Ph.D. degree with the Thrust of Robotics and Autonomous Systems, The Hong Kong University of Science and Technology (Guangzhou), Guangzhou, China. His research interests include autonomous navigation and robotics.
\end{IEEEbiography}
\vspace{-33pt}
\begin{IEEEbiography}[{\includegraphics[width=1in,height=1.25in,clip,keepaspectratio]{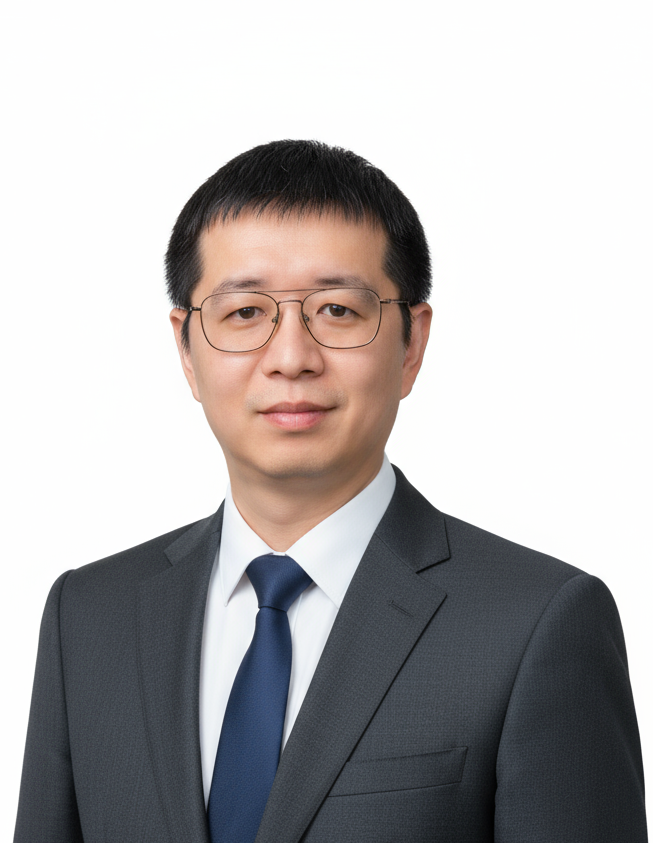}}]
{Xiao Hu} received a B.S. degree in Measurement and Control Technology from Northwestern Polytechnical University, Xian, China, in 2010. He later earned an M.S. degree in Computer and Communication Engineering from Universit\'e de Technologie de Compi\`egne, and finally, a Ph.D. degree from the Technical University of Denmark in 2021. 
He is currently a chief scientist at Astrall Dynamics Technology. Previously, he was a senior researcher at the Lower Airspace Economy Research Institute (LASER) within the International Digital Economy Academy (IDEA) and an Algorithm Expert at Alibaba DAMO Academy. His research interests primarily focus on computer vision and navigation for unmanned aerial vehicles.
\end{IEEEbiography}
\vspace{-33pt}
\begin{IEEEbiography}[{\includegraphics[width=1in,height=1.25in,clip,keepaspectratio]{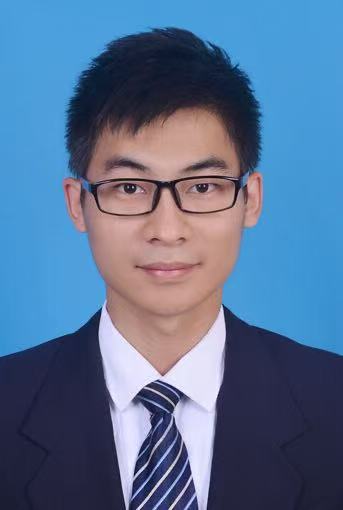}}]
{Wancheng  Liu} obtained a Master's degree in Electronics and Communication Engineering from Dalian University of Technology in 2016. He formerly served as an Algorithm Expert at Alibaba DAMO Academy and currently works as an Algorithm Engineer at Horizon-Continental Technology Corporation, where his primary focus is on the development of sensor calibration algorithms for autonomous driving.
\end{IEEEbiography}
\vspace{-33pt}
\begin{IEEEbiography}[{\includegraphics[width=1in,height=1.25in,clip,keepaspectratio]{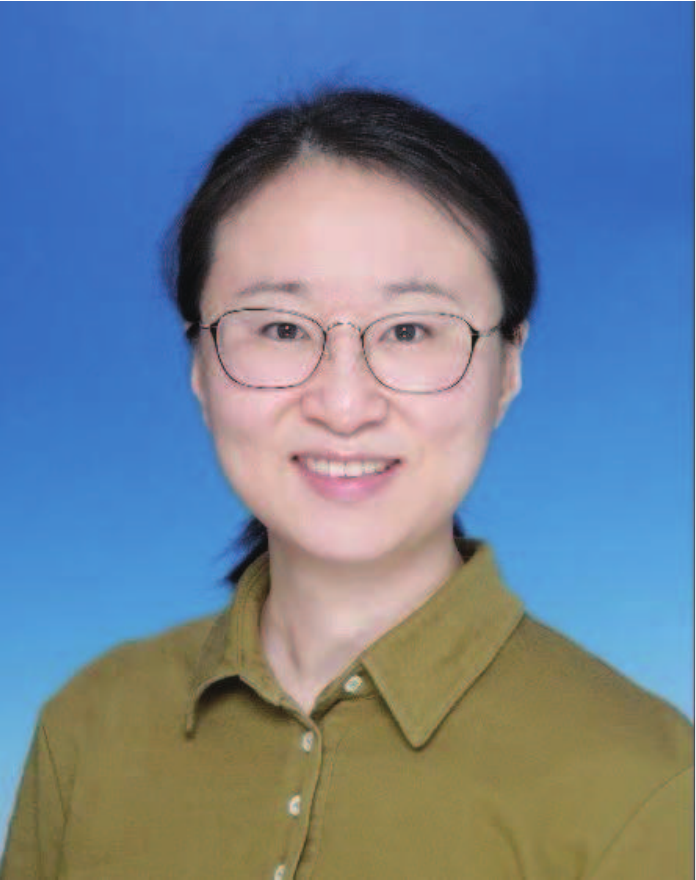}}]
{Wei Jiang} received the Ph.D. degree from the School of Civil and Environmental Engineering, University of New South Wales (UNSW), Sydney, Australia in 2015. 

She is a Professor with the School of Electronic and Information Engineering, Beijing Jiaotong University, Beijing, China. Her current research interest focuses on multi-sensor integration, and in particular the implementation of new navigation and data fusion algorithms.
\end{IEEEbiography}

\vfill

\end{document}